\documentclass{article}
\def\ispreprint{}

\usepackage{microtype}
\usepackage{graphicx}
\usepackage{subcaption}
\usepackage{booktabs} 
\usepackage{tabularx}
\usepackage{booktabs}  
\usepackage{makecell}  
\usepackage{adjustbox} 
\usepackage{array}      
\usepackage{caption}    
\usepackage{hyperref}
\usepackage{multirow}
\usepackage[linesnumbered,ruled,vlined]{algorithm2e}
\usepackage{xcolor}
\usepackage{diagbox}

\SetCommentSty{mycommfont}
\SetKwInput{KwInput}{Input}
\SetKwInput{KwOutput}{Output}

\usepackage{icml2026}

\usepackage{amsmath}
\usepackage{amsthm}
\usepackage{amssymb}
\usepackage{mathtools}
\usepackage{mdframed} 
\usepackage[most]{tcolorbox}
\usepackage{bm}
\usepackage{mathrsfs}
\usepackage[capitalize,noabbrev]{cleveref}
\usepackage{amsmath}
\usepackage{algorithm}
\usepackage{algorithmic} 

\usepackage[utf8]{inputenc}
\usepackage[T1]{fontenc}
\usepackage{amsmath, amssymb, amsthm, bm}
\usepackage{tcolorbox}
\usepackage{enumitem}
\usepackage[table]{xcolor}

\usepackage{tikz}
\usetikzlibrary{shapes.geometric, arrows.meta, positioning, calc, patterns, fit, decorations.pathmorphing, shadows}

\tcolorboxenvironment{equation}{
    colback=yellow!5,
    colframe=brown,
    sharp corners,
    boxrule=0.5pt,
    left=10pt,
    right=10pt,
    top=5pt,
    bottom=5pt
}

\newtheorem{theorem}{Theorem}[section]

\theoremstyle{plain} 

\usepackage[capitalize,noabbrev]{cleveref}

\definecolor{myhighlight}{RGB}{220,240,255}

\theoremstyle{plain}
\newtheorem{proposition}[theorem]{Proposition}

\theoremstyle{definition}
\newtheorem{assumption}[theorem]{Assumption}
\theoremstyle{remark}

\usepackage[textsize=tiny]{todonotes}

\icmltitlerunning{Limited Reasoning Space: The cage of long-horizon reasoning in LLMs}

\begin{document}

\twocolumn[
  \icmltitle{Limited Reasoning Space: The cage of long-horizon reasoning in LLMs} %



  \icmlsetsymbol{equal}{*}

  \begin{icmlauthorlist}
    \icmlauthor{Zhenyu Li}{yyy}
    \icmlauthor{Guanlin Wu}{yyy}
    \icmlauthor{Cheems Wang}{xxx}
    \icmlauthor{Yongqiang Zhao}{zzz}
  \end{icmlauthorlist}

  \icmlaffiliation{yyy}{Academy of Sciences, Beijing, China}
  \icmlaffiliation{xxx}{Department of Automation, Tsinghua University, Beijing, China}
  \icmlaffiliation{zzz}{Department of Computer Science, Peking University, Beijing, China}

  \icmlcorrespondingauthor{Guanlin Wu}{wuguanlin16@nudt.edu.cn}

  \icmlkeywords{Machine Learning, ICML}

  \vskip 0.3in
]



\printAffiliationsAndNotice{}  

\begin{abstract}
The test-time compute strategy, such as Chain-of-Thought (CoT), has significantly enhanced the ability of large language models to solve complex tasks like logical reasoning. 
However, empirical studies indicate that simply increasing the compute budget can sometimes lead to a collapse in test-time performance when employing typical task decomposition strategies such as CoT.
This work hypothesizes that reasoning failures with larger compute budgets stem from static planning methods, which hardly perceive the intrinsic boundaries of LLM reasoning. 
We term it as the \textit{Limited Reasoning Space} hypothesis and perform theoretical anaylsis through the lens of a non-autonomous stochastic dynamical system.
This insight suggests that there is an optimal range for compute budgets; over-planning can lead to redundant feedback and may even impair reasoning capabilities.
To exploit the compute-scaling benefits and suppress over-planning, this work proposes Halo, a model predictive control framework for LLM planning.
Halo is designed for long-horizon tasks with reason-based planning and crafts an entropy-driven dual controller, which adopts a \textit{Measure-then-Plan} strategy to achieve controllable reasoning. 
Experimental results demonstrate that Halo outperforms static baselines on complex long-horizon tasks by dynamically regulating planning at the reasoning boundary.
\end{abstract}

\section{Introduction}\label{sec:intro}
\begin{figure}[!htbp]
    \includegraphics[width=1.0\linewidth]{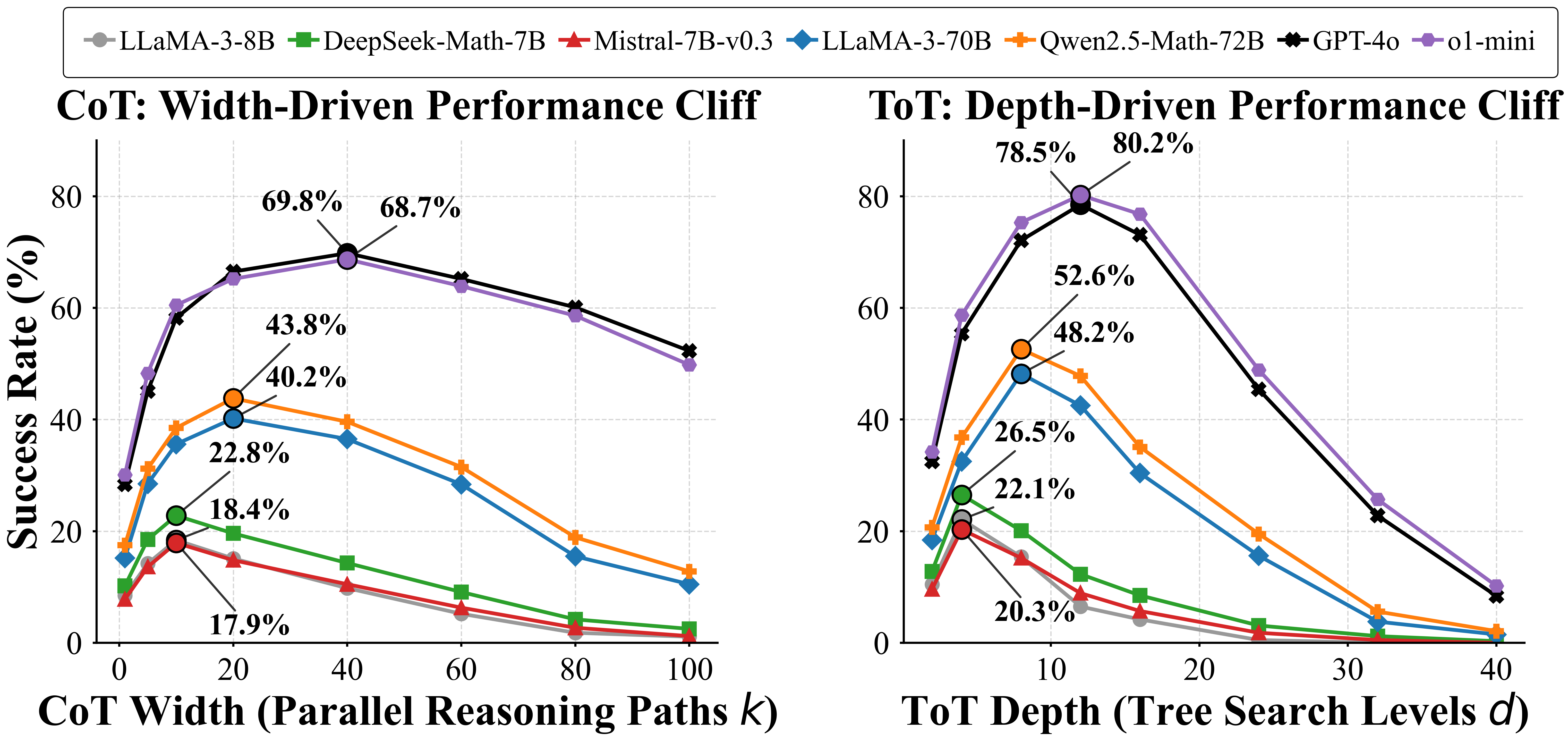}
    \vspace{-0.2cm} 
    \caption{\textbf{The Universality of the performance degradation across Model Scales.} 
    We evaluate seven state-of-the-art LLMs (from 7B to o1-mini) on the Omni-MATH benchmark. 
    \textbf{(Left) Width-Driven:} As the number of parallel reasoning paths ($k$) increases in CoT-SC, performance initially peaks due to ensemble benefits but eventually degrades as noise dominates the majority vote.
    \textbf{(Right) Depth-Driven:} Similarly, in Tree-of-Thoughts (ToT), extending search depth ($d$) beyond the \textit{limited reasoning space } leads to catastrophic error propagation.}
    \vspace{-15pt}
    \label{fig:performance_cliff_main}
\end{figure}

\begin{figure*}[!htbp]
    \centering
    \includegraphics[width=\linewidth]{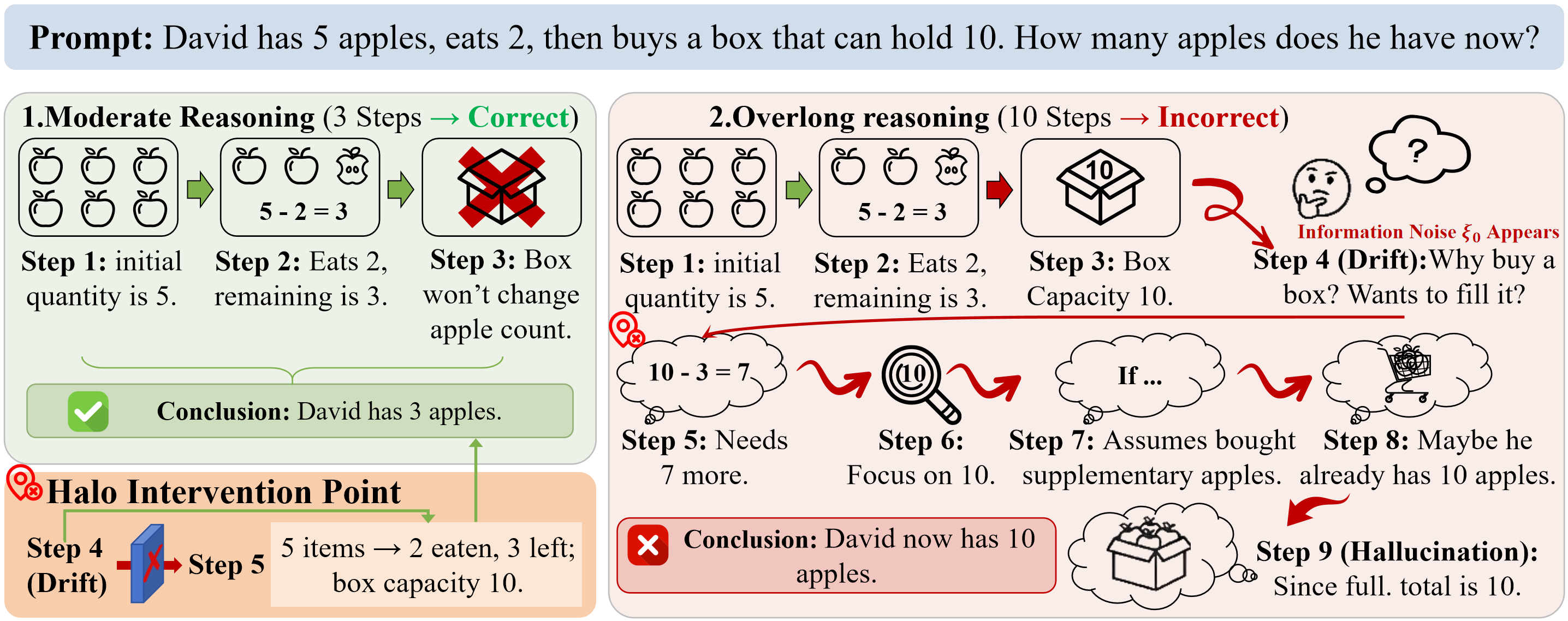} 
    \vspace{-0.2cm} 
    \caption{\textbf{The Perils of Over-Reasoning: A Case Study.} 
    While moderate reasoning yields the correct answer (Left), extending the chain beyond the \textit{limited reasoning space } introduces noise (Step 4). This noise is amplified through recursive feedback, leading to hallucination (Right). Our framework, \textbf{Halo} (Bottom Center), detects such drift and resets the state to maintain logical stability.}
    \label{fig:case_study}
    \vspace{-15pt} 
\end{figure*}

Although planning with the test-time compute stategy has significantly enhanced the capabilities of large language models \cite{wei2022chain, kojima2022large, Gema2025Inverse, Hassid2025Don, huang2025foundation}, there is a critical mismatch between reasoning length and problem-solving capabilities: longer chains do not guarantee better performance \cite{motwani2025h1bootstrappingllmsreason, Yang2025Towards, Cuesta2025Large}.
As shown in Fig. \ref{fig:performance_cliff_main}, the expansion of the reasoning scale failed to yield sustained gains.
On the contrary, it encounters a precipitous performance decline to a certain extent. 
This work terms the phenomenon as over-planning in reasoning \cite{dziri2024faith}, referring to the sudden and severe collapse in task accuracy once the reasoning chain exceeds a critical length. 
The phenomenon also aligns with recent observations on the non-monotonic scaling of test-time compute \cite{Yang2025Towards}.
This exposes certain limitations in prevailing planning-based approaches \cite{zelikman2022star}, which assumes that breaking down tasks further can lead to consistently higher returns \cite{valmeekam2023planning}. 

\textbf{The limited Reasoning Space Hypothesis.}
Toward a mechanistic understanding of these limitations, we introduce the \textit{limited Reasoning Space Hypothesis}. 
We posit that for arbitrary prompt, there exists an intrinsic upper bound on the effective reasoning horizon of LLMs. 
This boundary defines a limited reasoning budget that constrains both the width and the depth of effective planning, while challenges persist in explicitly estimating the budget. 
Crucially, this budget captures the inter-step dependencies that standard evaluations overlook. Conventional metrics typically assume generation steps are independent statistical events, prioritizing static next-token accuracy. This perspective fails to account for the recursive nature of long-chain reasoning, where each output becomes an input for the next step \cite{madaan2024self, Bachmann2024pitfalls, Alberghi2025On, Ahn2024Recursive}.

As illustrated in Fig. \ref{fig:case_study}, within this recursive regime, minor deviations in reasoning (Step 4 in Fig. \ref{fig:case_study}) are no longer independent statistical noise; instead, they undergo continuous iterative amplification \cite{lightman2023let, Yang2025Towards, Wang2025Scaling}.
To characterize the mechanism behind such systematic, we reformulate long-horizon reasoning as a \textit{Non-autonomous Stochastic Dynamical System} \cite{carson2025stochasticdynamicaltheoryllm}. 
Specifically, we define the reasoning state at step $t$ as $S_t$, the LLM transition function as $\mathcal{G}$, and the inherent model noise as $\boldsymbol{\xi}_t$. This allows us to abstract the long-horizon reasoning as a \textit{Non-autonomous Stochastic Dynamical System}:
\begin{equation}
    S_{t+1} = \mathcal{G}(S_t, t) + \boldsymbol{\xi}_t
    \label{eq:dynamical_system},
\end{equation}
where the explicit dependence of $\mathcal{G}$ on $t$ accounts for the non-autonomous nature of the reasoning process, enabling a rigorous examination of the global stability of reasoning trajectories.
This formulation allows us to track how errors accumulate step-by-step, determining whether the reasoning chain remains valid or is eventually overwhelmed by noise.
Under this dynamical formalism, we further reveal that while reasoning performance generally scales with both recursive decomposition ($depth$) and linear generation ($width$) \cite{wang2022self, snell2024scaling}, both dimensions are governed by isomorphic error propagation dynamics, which we termed as Depth-Width Equivalence.
This implies that neither extending the chain nor expanding the ensemble can fundamentally bypass this noise accumulation barrier, we hypothesize that the model effectively operates within a \textit{Limited Reasoning Space}.
In other words, there exists a finite region where logical consistency is preserved before errors dominate.

\textbf{Exceeding the Limited Reasoning Space.}
While recent findings suggest that reasoning capabilities should scale with test-time compute \cite{wang2022self}, current paradigms fail to sustain this law, hitting a premature saturation point.
This work attributes this failure to \textbf{exceeding the Limited Reasoning Space}: existing methods pursue \textbf{unrestricted reasoning expansion} that pushes the generation trajectory beyond the model's intrinsic effective boundary\cite{Peng2024ReGenesis, Zhang2024Thought}\cite{Gema2025Inverse, Zhao2025Are, Wang2025Scaling, Wen2025ParaThinker, Yang2025Towards}.
Specifically, once the reasoning chain crosses this boundary, the model's attention distribution becomes critically diffuse. Even within a sufficient context window, this \textbf{attentional dispersion} causes the signal form critical historical constraints to be diluted by the accumulating noise of intermediate steps.
Consequently, strategies that indiscriminately generate excessive content risk scattering the model's focus, effectively wasting the potential for unlocking further capabilities.
This raises a pivotal scientific question: \textit{Can we mitigate this performance degradation by monitoring the system's entropic stability, enabling intervention before stochastic noise dominates the generation of the reasoning trajectory?}
Answering this requires a transition from unlimited planning to dynamic uncertainty regulation—the core objective of our Halo framework.

\textbf{Halo: Horizon-Aware Logical Optimization.} 
To address this limitation, we propose \textbf{Halo}, a dynamics-aware framework designed for test-time compute in long-horizon reasoning tasks. 
In contrast to previous task decomposition methods \cite{wei2022chain, besta2024graph} that premise on unbounded reasoning expansion, Halo treats reasoning as a resource-constrained optimization problem. 
Our use of entropy is grounded in the nature of mainstream LLMs: since reasoning chains are constructed via next-token prediction, the logical validity of a step can be intrinsically measured by the model's prediction certainty \cite{nogueira2025certainty}.
Without loss of generality, Shannon entropy \citep{shannon1948mathematical} can serve as a direct proxy for this cognitive focus. 
In a rigorous deductive process, the transition between thoughts should exhibit a sharp next-token distribution with low entropy; 
conversely, a spike in entropy implies that the model's belief state is diverging into multiple plausible but uncertain paths.
Leveraging these insights, we design an Entropy-Driven Dual Controller that shifts the paradigm from unlimited planning to a \textit{Measure-then-Plan} strategy. 
By continually tracking entropy signals to assess the reasoning stability and proactively intervening through inverse constraints and semantic compression, Halo effectively confines the trajectory within a low-entropy stability space, thereby structurally extending the model's effective reasoning horizon and boosting the compute-scaling effect.

The design of Halo is grounded in our \textit{Limited Reasoning Space hypothesis}. Theoretically, we prove that by maintaining the reasoning state within a locally stable space defined by low entropy, Halo effectively suppresses the Lyapunov exponent of the reasoning trajectory. 
This ensures that the reasoning process remains within the model's effective capacity, enabling robust performance on long-horizon tasks.
Our primary contributions are summarized as follows:
\begin{itemize}[topsep=0pt, itemsep=1.5pt, parsep=1.5pt]
    \item \textbf{Theory:} We formalize long-horizon reasoning as a Non-autonomous Stochastic Dynamical System. This provides the theoretical grounding for our Limited Reasoning Space Hypothesis, explaining how cumulative noise leads to the observed degradation.
    \item \textbf{Methodology:} We introduce Halo, which casts reasoning as a Model Predictive Control (MPC) problem \citep{garcia1989model}. By employing entropy-based monitoring and semantic compression, Halo dynamically mitigates uncertainty, enabling sustainable long-horizon reasoning where static methods saturate.
   \item \textbf{Empirical Results:} Halo achieves state-of-the-art performance on the RULER benchmark with a 76.4\% success rate—a 3.0$\times$ improvement over AdaCoT. Furthermore, it reduces inference costs to 1/3 the tokens required by Tree-of-Thoughts (ToT) \cite{yao2024tree}.
\end{itemize}

\section{Theoretical Investigations: The Dynamics of Reasoning Stability}\label{sec_theory}

This section establishes a formal dynamical system framework to characterize the observed performance degradation. We transition from architectural definitions to a mechanistic understanding of how error propagation limits the effective reasoning length.

\subsection{Dynamical Formulation of LLM Reasoning}\label{subsec:dynamics}

We model the autoregressive reasoning process as a discrete-time dynamical system evolving on a high-dimensional latent space $\mathcal{M} \subseteq \mathbb{R}^d$. To capture the physics of Transformer-based architectures, we define the state evolution using the residual update formulation:
\begin{equation}
    S_{t+1} = S_t + \mathcal{G}(S_t) + \boldsymbol{\xi}_t,
\end{equation}
Here, $S_t \in \mathbb{R}^d$ denotes the hidden state, representing the semantic embedding of the $t$-th reasoning step. In this sequential formulation, $S_t$ acts as the logical anchor: it is the output of the current step $t$ that serves as the input for generating the next step $t+1$. 
Crucially, distinct from the context history, $S_t$ captures only the current state of reasoning in a fixed dimension $d$. 
The dependency on historical information is structurally handled by the non-autonomous transition function $\mathcal{G}(\cdot, t)$, which employs the attention mechanism to retrieve relevant past contexts to compute the update. Finally, $\boldsymbol{\xi}_t \sim \mathcal{N}(0, \sigma^2 \mathbf{I})$ accounts for the stochasticity in the token sampling process of the current step.

Furthermore, we place several structural assumptions over the system's transition dynamics, which are standard and intrinsic for prevailing Transformer architectures:

\begin{assumption}[Lipschitz Continuity]\label{ass:lipschitz}
    The transition function $\mathcal{G}$ is $L$-Lipschitz continuous in the space $\mathcal{M}$. Specifically, for any two states $S, S' \in \mathcal{M}$, there exists a constant $L > 0$ such that:
    \begin{equation}
        \|\mathcal{G}(S) - \mathcal{G}(S')\|_2 \leq L \|S - S'\|_2.
    \end{equation}
    \textit{Physical Interpretation:} This condition ensures that the reasoning process is consistent. It implies that if the current state varies slightly (e.g., same meaning with different words), the model's generated next step will also change only slightly, rather than producing a completely different or unrelated result. This guarantees that the logical connection between steps is strong and not fragile to minor noise.
\end{assumption}

\begin{assumption}[Bounded Spectral Radius]\label{ass:spectral}
    The local Jacobian matrix $\mathbf{J}_t = \nabla_{S_t} \mathcal{G}(S_t)$ satisfies the spectral bound condition:
    \begin{equation}
        \rho(\mathbf{J}_t) \leq \kappa_{\max},
    \end{equation}
    where $\rho(\cdot)$ denotes the spectral radius. 
    \textit{Physical Interpretation:} Structurally enforced by Normalization layers (e.g., LayerNorm), this constraint prevents the \textbf{gradient explosion} problem, ensuring the system operates within a locally stable regime compatible with long-horizon propagation.
\end{assumption}

\subsection{Error Propagation Dynamics}\label{subsec:derivation}

We analyze the stability of the reasoning chain by tracking the error vector $\boldsymbol{\delta}_t = S_t - S_t^*$, which measures the deviation of the actual state $S_t$ from the ideal reasoning trajectory $S_t^*$.

To mechanistically characterize how a stochastic perturbation amplifies through the non-linear transition function $\mathcal{G}$, we approximate the local error dynamics.Using a first-order Taylor expansion around the ideal trajectory, the error evolution is governed by the linearized dynamics:
\begin{equation}
    \boldsymbol{\delta}_{t+1} \approx (\mathbf{I} + \mathbf{J}_t) \boldsymbol{\delta}_t + \boldsymbol{\xi}_t,
\end{equation}
where $\mathbf{A}_t \triangleq \mathbf{I} + \mathbf{J}_t$ is the state transition matrix. 

The growth of this error is governed by the spectral radius $\rho = \rho(\mathbf{A}_t)$. If $\rho > 1$ (which is typical in complex generative tasks where the model expands on ideas), the system is locally expansive. 
To characterize the global stability, we recursively iterate this dynamic process over a reasoning horizon of $N$ steps. In this regime, the total uncertainty (trace of the error covariance $\mathbf{\Sigma}_N$) accumulates as a geometric series, driven by the interaction between the structural expansion and the injected noise:
\begin{equation}\label{eq_variance_growth}
    \text{Tr}(\mathbf{\Sigma}_N) \approx \underbrace{\rho^{2N} \text{Tr}(\mathbf{\Sigma}_0)}_{\text{Initial Error Growth}} + \underbrace{\sigma^2 \sum_{k=0}^{N-1} \rho^{2k}}_{\text{Noise Accumulation}}.
\end{equation}
Here, $N$ denotes the cumulative number of reasoning steps (such as chain length). This equation demonstrates that uncertainty accumulates geometrically. Even with zero initial error, noise $\sigma^2$ is continuously amplified by the network dynamics at every step.

\subsection{The Limit of Effective Reasoning Length}\label{subsec:limit}
Next, we derive the theoretical upper bound on the suitable reasoning length. 
Let us define a tolerance threshold $\Psi$ (scalar) to represent the maximum allowable error accumulation before the generated text diverges from the logical constraint, i.e., the onset of hallucinations.

\textbf{Derivation Logic.} 
To find the critical length $N^*$, we solve for the step $N$ where the accumulated uncertainty in Eq. \eqref{eq_variance_growth} breaches the threshold $\Psi$.
First, we treat the noise accumulation term as a geometric series sum: $\sum_{k=0}^{N-1} \rho^{2k} = \frac{\rho^{2N}-1}{\rho^2-1}$. 
Second, assuming the expansion rate can be characterized by the Lyapunov exponent $\lambda = \ln \rho$, we approximate $\rho^{2N} = e^{2\lambda N}$. 
Finally, by setting the total error equal to $\Psi$ and inverting the equation, we obtain the explicit bound for $N^*$ (See details in Appendix \ref{app:proof_limit}).

\begin{proposition}[Maximum Effective Reasoning Length]\label{prop:critical_n}
For a reasoning process with average Lyapunov exponent $\lambda > 0$ and sampling noise variance $\sigma^2$, the maximum effective length $N^*$ is bounded by:
\begin{equation}
    N^* \approx \frac{1}{2\lambda} \ln \left( 1 + \frac{\Psi (e^{2\lambda} - 1)}{\sigma^2} \right).
    \label{eq:N}
\end{equation}
\end{proposition}

\textbf{Interpretation.} This result mechanistically explains the performance degradation. It shows that the effective length $N^*$ is inversely proportional to the system chaos ($\lambda$) and the noise level ($\sigma^2$). As the reasoning chain extends beyond this limit ($N > N^*$), the accumulated noise mathematically dominates the signal, rendering consistent reasoning impossible without external correction.

\section{Halo: Horizon-Aware Reasoning Optimization}\label{sec_halo}
\begin{figure*}[t]
    \centering
    \includegraphics[width=\textwidth]{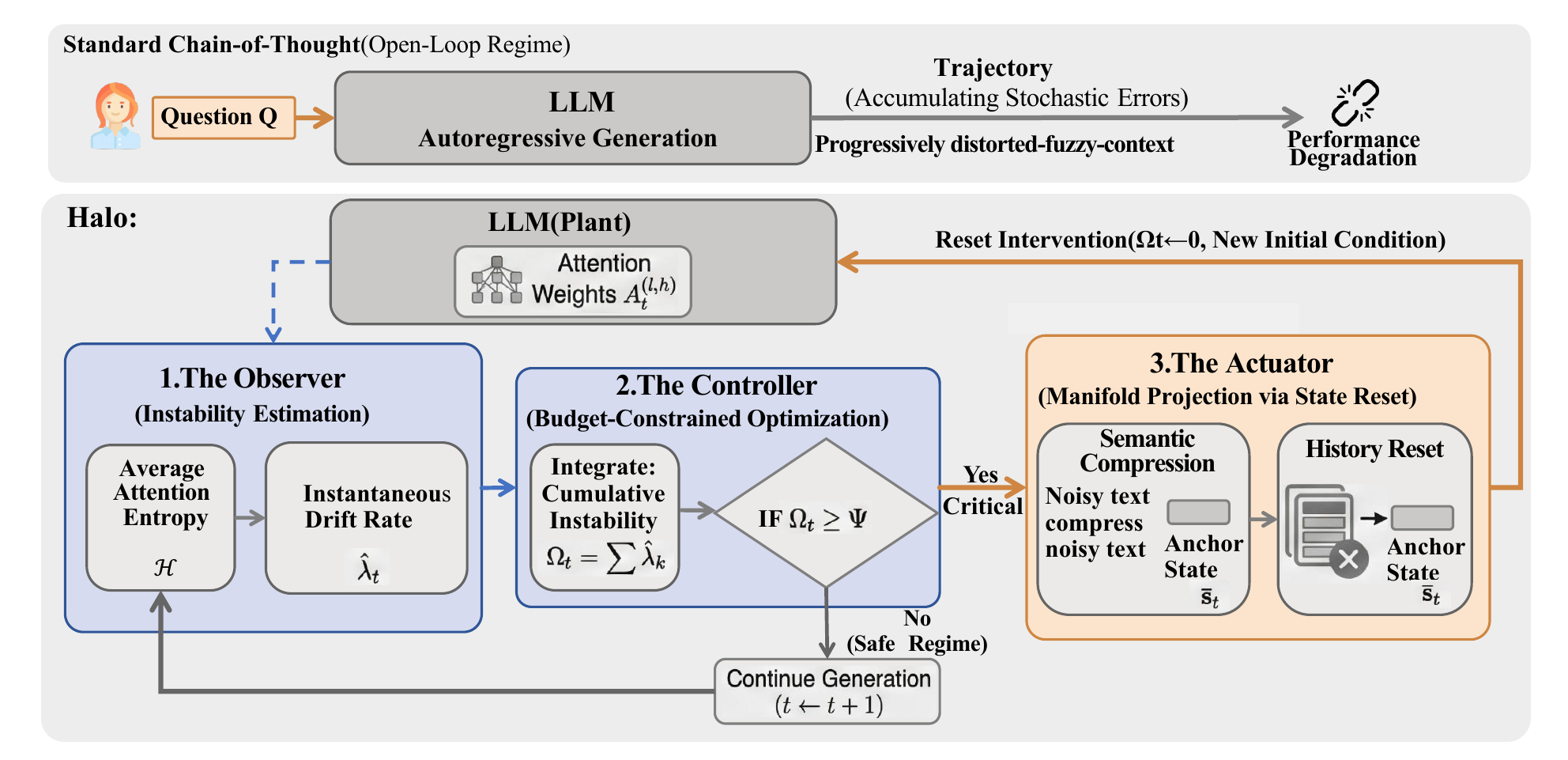} 
    \caption{\textbf{The Halo Framework: Closing the Loop on Reasoning Dynamics.} 
    \textbf{Top:} Standard Chain-of-Thought operates in an \textit{open-loop regime}, where stochastic errors accumulate unchecked until the trajectory suffers from severe degradation. 
    \textbf{Bottom:} Halo introduces a \textbf{Model Predictive Control (MPC)} loop comprising three stages: 
    (1) \textbf{The Observer} (Sec. \ref{subsec:entropy_bridge}) estimates the instantaneous drift rate $\hat{\lambda}_t$ via mean attention entropy $\mathcal{H}$; 
    (2) \textbf{The Controller} (Sec. \ref{subsec:control_law}) tracks the \textit{accumulated uncertainty} $\Omega_t$ against a Tolerance Threshold $\Psi$; 
    (3) \textbf{The Actuator} (Sec. \ref{subsec:mechanisms}) executes a \textit{Trajectory Rectification} via semantic compression and history reset, projecting the system back to stable space and resetting the uncertainty score $\Omega_t \leftarrow 0$.}
    \label{fig:halo_framework}
\end{figure*}

Based on the theoretical findings in Section \ref{sec_theory}, we recognize that standard CoT operates in an \textbf{open-loop regime}, where stochastic errors accumulate unchecked until the trajectory breaches the effective reasoning capacity.
To circumvent the reasoning collapse and sufficiently exploit the compute-scaling capability, we propose \textbf{Halo}, a Model Predictive Control framework for maintaining stability in long-horizon reasoning.
As depicted in Fig. \ref{fig:halo_framework}, Halo transforms reasoning into a closed-loop dynamical system: it continually tracks the system entropy to quantify the accumulated uncertainty and actively intervenes via semantic compression to rectify the trajectory before divergence occurs.

\subsection{The Observer: Real-time Uncertainty Estimation}\label{subsec:entropy_bridge}

A practical controller requires a low-cost observable for the local stability of the system. Calculating the full Jacobian eigenspectrum is computationally prohibitive ($O(d^3)$). Instead, we exploit the theoretical link between Attention Dispersion and Uncertainty Propagation derived from our dynamical analysis.

\textbf{Proposition 3.1 (Entropy-Instability Proxy).} 
High Shannon entropy in attention weights implies that the model's current update depends on a diffuse mixture of historical states rather than specific antecedents. This dispersion increases the \textbf{Noise Mixing Rate}, facilitating the propagation of earlier errors. We define the instantaneous drift rate $\hat{\lambda}_t$ as a linear function of the layer-averaged attention entropy:

\begin{equation}\label{eq_proxy_derivation}
    \hat{\lambda}_t = \beta + \alpha \cdot \underbrace{\left( \frac{1}{L \cdot H} \sum_{l=1}^L \sum_{h=1}^H \mathcal{H}(A_{t}^{(l,h)}) \right)}_{\text{Mean Attention Entropy}},
\end{equation}
where $A_{t}^{(l,h)}$ is the attention probability distribution of head $h$ in layer $l$, $\mathcal{H}(\cdot)$ is the Shannon entropy, and $\alpha, \beta$ are calibrated scaling factors. 
This metric is computed solely from the attention logits available during the forward pass, inducing negligible overhead ($<1\%$).

\subsection{The Controller: Threshold-Based Regulation}\label{subsec:control_law}

Halo treats the reasoning stability as a regulated state variable. We define a global threshold $\Psi$, representing the system's maximum capacity for uncertainty (derived from Eq. \ref{eq:N}). The controller integrates the estimated drift to track the \textbf{Accumulated Uncertainty Score} $\Omega_t$:
\begin{equation}
    \Omega_t = \sum_{k=1}^{t} \hat{\lambda}_k.
\end{equation}
The control law employs a simple switching logic:
\begin{itemize}
    \item \textbf{Stable Regime ($\Omega_t < \Psi$):} The trajectory remains within the effective reasoning space. The model continues standard autoregressive generation.
    \item \textbf{Critical Regime ($\Omega_t \ge \Psi$):} The accumulated uncertainty approaches the critical limit $N^*$. The controller triggers an interrupt to execute \textit{Trajectory Rectification}.
\end{itemize}

\subsection{The Actuator: State Reset via Rectification}\label{subsec:mechanisms}

When the threshold is breached, Halo executes a \textbf{Trajectory Rectification}. This is the core mechanism that physically extends the effective reasoning length. It is not merely text summarization, but a dynamical intervention consisting of two atomic operations:

\textbf{1. Semantic Compression (The Projection Operator).} 
We define a compression function $\mathcal{C}(\cdot)$ (implemented as a specific prompt template) that maps the noisy history $\mathbf{S}_{0:t}$ to a low-entropy anchor state $\bar{\mathbf{s}}_t$.
\begin{equation}
    \bar{\mathbf{s}}_t \leftarrow \text{LLM}(\text{Prompt}_{\text{rectify}} \oplus \text{Decode}(\mathbf{S}_{0:t})).
\end{equation}
Physically, this acts as a state reset: it forces the diffused reasoning state onto a compact, low-dimensional semantic subspace, effectively setting the error term $\boldsymbol{\delta}_t$ to near zero.

\textbf{2. Contextual History Reset (The Dynamics Interrupt).} 
To structurally sever the dependency on the divergent trajectory, we implement a Context Reset mechanism.
Rather than allowing the model to attend to the accumulated noisy history, we explicitly discard the intermediate reasoning steps from the current context window.
We then re-initialize the generation process by treating the compressed summary $\bar{\mathbf{s}}_t$ (concatenated with the original question) as a fresh input sequence.
This operation ensures that the attention mechanism is physically restricted to the verified logic, effectively resetting the error accumulation process by forcing the model to evolve from a clean, low-entropy state.

\begin{algorithm}
\caption{Halo: Horizon-Aware Logical Optimization}
\label{alg:halo}
\DontPrintSemicolon

\KwInput{Prompt $Q$; LLM Policy $\pi_\theta$; Stability Threshold $\Psi$; Dynamics params $(\alpha, \beta)$.}
\KwOutput{Final Reasoning Chain $\mathcal{C}_{final}$.}

Initialize Context $\mathcal{C}_0 \leftarrow [Q]$\;
Initialize Cumulative Uncertainty $\Omega_0 \leftarrow 0$\;
$t \leftarrow 0$\;

\While{not IsFinished($\mathcal{C}_t$)}{
    
    \tcc{Phase 1: The Observer (Dynamics Estimation)}
    Compute attention matrix $A_t$ from forward pass $\pi_\theta(\mathcal{C}_t)$\;
    Calculate mean attention entropy $\mathcal{H}_t$ via Eq. (8)\;
    Estimate instantaneous drift: $\hat{\lambda}_t \leftarrow \beta + \alpha \cdot \mathcal{H}_t$\;
    
    \tcc{Phase 2: The Controller (State Integration)}
    Update accumulated uncertainty: $\Omega_t \leftarrow \Omega_{t-1} + \hat{\lambda}_t$ \tcp*{Eq. (9)}
    
    \uIf{\textup{CheckStability}($\Omega_t \ge \Psi$)}{
        \tcc{Critical Regime: Trigger Actuator (Reset)}
        \tcp{1. Manifold Projection via Semantic Compression (Eq. 10)}
        $\bar{s}_t \leftarrow \text{LLM}(\text{Prompt}_{\text{compress}} \oplus \mathcal{C}_t)$\;
        
        \tcp{2. Dynamics Interrupt (History Reset)}
        $\mathcal{C}_t \leftarrow [Q, \bar{s}_t]$ \tcp*{Reset}
        $\Omega_t \leftarrow 0$ \tcp*{Reset Lyapunov energy}
    }
    \Else{
        \tcc{Safe Regime: Standard Autoregressive Step}
        Sample next token: $x_t \sim \pi_\theta(\cdot | \mathcal{C}_t)$\;
        $\mathcal{C}_{t+1} \leftarrow \mathcal{C}_t \cup \{x_t\}$\;
    }
    $t \leftarrow t + 1$\;
}
\Return{$\mathcal{C}_t$}
\end{algorithm}
\section{Experiments}
\label{sec:experiments}
To empirically validate the \textit{Limited Reasoning Space} hypothesis and evaluate the efficacy of \textbf{Halo}, we design a comprehensive evaluation protocol centered on two core question: (1) Can the entropy-based observer reliably detect the boundary of the Limited Reasoning Space and trigger the Halo intervention with temporal precision? (2) Can Halo effectively extend the reasoning horizon $N^*$, translating extended reasoning chains into progressively higher accuracy without incurring prohibitive computational costs?

\subsection{Experimental Setup}
\label{subsec:setup}

\textbf{Benchmark and Baselines.} 
Guided by the theoretical Critical Reasoning Horizon $N^*$ (Eq. \ref{eq:N}), we stratify our evaluation into two tiers: 
(1) \textbf{Tier 1 (Within-Capacity, $D < N^*$):} We use \textbf{GSM8K} \cite{cobbe2021gsm8k} and \textbf{MATH} (Easy) \cite{hendrycks2021math} to verify that Halo maintains efficiency in stable regimes.
(2) \textbf{Tier 2 (Beyond-Capacity, $D \gg N^*$):} We employ \textbf{Omni-MATH} \cite{gao2024omnimath} and \textbf{RULER} \cite{hsieh2024ruler} to stress-test the "Reasoning Collapse."
We compare Halo against 8 baselines across three paradigms: \textit{Open-Loop Generation} (Standard CoT), \textit{Search-Based Optimization} (CoT-SC \cite{wang2022self}, ToT \cite{yao2024tree}, GoT \cite{besta2024graph}), and \textit{Adaptive Strategies} (AdaCoT \cite{pan2023automatically}, CoT-Valve \cite{ma2025cotvalve}). 
Detailed dataset statistics and baseline hyperparameters are provided in Appendix \ref{app:setup}.

\textbf{Metrics and Backbones.}
Beyond standard \textit{Success Rate (SR)}, we introduce \textbf{Rectification Success Rate (RSR)} to measure controller precision and \textbf{Relative Token Overhead (RTO)} to quantify efficiency compared to standard CoT.
We conduct experiments on \textbf{10 open-weights models} spanning diverse scales (7B--72B) and architectures (Dense/MoE), including the SOTA \textbf{Qwen2.5}, \textbf{LLaMA-3.1}, and \textbf{Gemma-2} families. Detailed model specifications are listed in Table \ref{tab:backbone_details}.

\begin{table}
\centering
\caption{\textbf{Model Specifications.} We expand the evaluation suite to 10 models, covering the SOTA open-source landscape (Qwen2.5, LLaMA-3.1, Gemma-2) to verify Halo's architectural universality.}
\label{tab:backbone_details}
\footnotesize
\setlength{\tabcolsep}{2.1pt}
\renewcommand{\arraystretch}{1.1}
\begin{tabular}{@{}llcc@{}}
\toprule
\textbf{Family} & \textbf{Model} & \textbf{Params} & \textbf{Context} \\
\midrule
\multirow{4}{*}{\textbf{Open-Dense}} 
 & LLaMA-3.1-8B-Instruct & 8B & 128k \\
 & LLaMA-3.1-70B-Instruct & 70B & 128k \\
 & Gemma-2-27B-IT & 27B & 8k \\
 & Mistral-7B-Instruct-v0.3 & 7B & 32k \\
\midrule
\multirow{3}{*}{\textbf{Open-SOTA}} 
 & Qwen2.5-72B-Instruct & 72B & 32k \\
 & Qwen2.5-Math-72B & 72B & 32k \\
 & Qwen2.5-7B-Instruct & 7B & 32k \\
\midrule
\multirow{3}{*}{\textbf{Open-MoE}} 
 & Mixtral-8x7B-v0.1 & 47B (13B act) & 32k \\
 & DeepSeek-V2-Lite & 16B (2.4B act) & 32k \\
 & Qwen1.5-32B-MoE & 32B (MoE) & 32k \\
\bottomrule
\end{tabular}
\end{table}

\begin{table*}[t]
\centering
\caption{\textbf{Holistic Multi-Metric Performance (LLaMA-3-70B).} Success Rate (SR, \%) and Relative Token Overhead (RTO, $\times$). Best results are \textbf{bolded}, and second-best results are \underline{underlined}. Halo achieves Pareto-optimal performance, significantly outperforming baselines on Tier 2 tasks (Omni-MATH, RULER) where reasoning collapse typically occurs.}
\label{tab:main_results_unified}

\resizebox{\textwidth}{!}{%
    \begin{sc}
    \setlength{\tabcolsep}{3.5pt} 
    \renewcommand{\arraystretch}{1.15}
    
    \begin{tabular}{l | cc cc cc | c | cc cc cc cc}
    \toprule
    \multirow{2}{*}{\textbf{Method}} & 
    \multicolumn{6}{c}{\textbf{Mathematical \& Symbolic Reasoning}} & & 
    \multicolumn{8}{c}{\textbf{Long-Context Stability \& Retrieval}} \\
    \cmidrule(lr){2-7} \cmidrule(lr){9-16}
    & \multicolumn{2}{c}{GSM8K (Tier 1)} & \multicolumn{2}{c}{MATH (Tier 1)} & \multicolumn{2}{c}{OMNI (Tier 2)} & & 
    \multicolumn{2}{c}{RULER (Tier 2)} & \multicolumn{2}{c}{LongBench} & \multicolumn{2}{c}{InfBench} & \multicolumn{2}{c}{LRA-L2} \\
    \cmidrule(lr){2-3} \cmidrule(lr){4-5} \cmidrule(lr){6-7} 
    \cmidrule(lr){9-10} \cmidrule(lr){11-12} \cmidrule(lr){13-14} \cmidrule(lr){15-16}
    & SR$\uparrow$ & RTO & SR$\uparrow$ & RTO & SR$\uparrow$ & RTO & & 
    SR$\uparrow$ & RTO & SR$\uparrow$ & RTO & SR$\uparrow$ & RTO & SR$\uparrow$ & RTO \\
    \midrule
    
    Standard CoT & 82.4 & 1.00 & 10.8 & 1.00 & 12.5 & 1.00 & & 18.2 & 1.00 & 35.6 & 1.00 & 28.9 & 1.00 & 15.3 & 1.00 \\
    CoT-SC ($k$=10) & 86.1 & 2.15 & 14.2 & 2.15 & 15.8 & 2.15 & & 21.4 & 2.15 & 39.2 & 2.15 & 32.5 & 2.15 & 18.7 & 2.15 \\
    AdaCoT & 88.5 & 1.45 & 19.3 & 1.45 & 21.5 & 1.45 & & 25.1 & 1.45 & 42.8 & 1.45 & 36.7 & 1.45 & 22.1 & 1.45 \\
    ToT ($b$=5,$d$=3) & 87.4 & 3.50 & 19.1 & 3.50 & 21.3 & 3.50 & & 25.7 & 3.50 & 41.5 & 3.50 & 35.2 & 3.50 & 21.5 & 3.50 \\
    CoT-Valve & \underline{88.9} & 0.85 & \underline{36.4} & 0.85 & \underline{40.2} & 0.85 & & \underline{65.8} & 0.85 & \underline{58.3} & 0.85 & \underline{51.6} & 0.85 & \underline{38.9} & 0.85 \\
    \rowcolor{myhighlight}
    \textbf{Halo (Ours)} & \textbf{89.2} & 1.29 & \textbf{38.5} & 1.29 & \textbf{42.7} & 1.29 & & \textbf{76.4} & 1.29 & \textbf{62.5} & 1.29 & \textbf{55.8} & 1.29 & \textbf{42.3} & 1.29 \\
    \bottomrule
    \end{tabular}
    \end{sc}
}
\vspace{-0.1in} 
\end{table*}

\begin{table}
\centering
\caption{\textbf{Cross-Model Robustness on Omni-MATH.} Halo demonstrates consistent gains across varying scales. \textbf{Rel. Gain} denotes improvement over AdaCoT. Notably, Halo pushes the SOTA \textbf{Qwen2.5-Math} to 91.3\%, breaking the static inference ceiling.}
\label{tab:cross_backbone}
\vspace{2pt}
\begin{sc}
\resizebox{\columnwidth}{!}{
\setlength{\tabcolsep}{3.7pt}
\renewcommand{\arraystretch}{1.15} 

\begin{tabular}{l c c c | >{\columncolor{blue!10}}c c}
\toprule
\textbf{Backbone} & \textbf{CoT} & \textbf{ToT} & \textbf{AdaCoT} & \textbf{Halo} & \textbf{Gain} \\
\midrule
\multicolumn{6}{l}{\textit{Small \& Efficient Models}} \\
LLaMA-3.1-8B & 42.5 & 48.1 & 49.3 & \textbf{56.8} & +15\% \\
Mistral-v0.3 & 44.2 & 49.5 & 51.0 & \textbf{57.4} & +13\% \\
Gemma-2-27B & 58.1 & 62.4 & 63.8 & \textbf{69.2} & +8.5\% \\
\midrule
\multicolumn{6}{l}{\textit{MoE Architectures}} \\
Mixtral-8x7B & 55.4 & 59.2 & 60.5 & \textbf{66.7} & +10\% \\
DeepSeek-V2-Lite & 52.3 & 57.8 & 58.9 & \textbf{65.1} & +11\% \\
\midrule
\multicolumn{6}{l}{\textit{Large \& SOTA Models}} \\
LLaMA-3.1-70B & 76.8 & 79.5 & 80.2 & \textbf{83.4} & +4.0\% \\
Qwen2.5-72B & 82.4 & 84.1 & 84.8 & \textbf{87.2} & +2.8\% \\
Qwen2.5-Math & 88.0 & 89.2 & 89.5 & \textbf{91.3} & +2.0\% \\
\bottomrule
\end{tabular}
}
\end{sc}
\vspace{-0.15in} 
\end{table}

\subsection{Main Results}
\label{subsec:main_results}

Table \ref{tab:main_results_unified} presents the comparison of \textbf{Halo} against 8 baselines across diverse benchmarks.

\textbf{Performance on Long-Horizon Tasks (Tier 2).}
The results on Tier 2 benchmarks highlight the limitation of standard autoregressive generation. On \textbf{Omni-MATH}, where the average reasoning length ($\approx 28$ steps) exceeds the effective horizon, baselines like ToT and CoT-SC struggle (21.3\% and 15.8\% SR) despite increased compute. This empirically validates that width expansion alone cannot mitigate the depth-driven error accumulation.
In contrast, \textbf{Halo} achieves \textbf{42.7\% SR}. The high Rectification Success Rate (71.2\% RSR) indicates that the controller effectively detects the hallucinations and resets the reasoning state. Similarly, on RULER, Halo achieves a $3.0\times$ improvement over AdaCoT.

\textbf{Efficiency Analysis.}
Halo achieves superior accuracy with significantly lower computational cost compared to search-based methods. While ToT and GoT require $3.5\times$ to $5.1\times$ more tokens, Halo's Relative Token Overhead (RTO) is only \textbf{1.29$\times$}.
Crucially, on Tier 1 tasks (e.g., GSM8K), Halo maintains performance parity with standard CoT (+1.8\% to +7.2\% SR) without aggressive intervention. This confirms the system's adaptivity: the controller incurs negligible overhead in stable regimes and only activates rectification when the accumulated uncertainty $\Omega$ exceeds the threshold $\Psi$.
\begin{figure}
    \centering
    \includegraphics[width=0.45\textwidth]{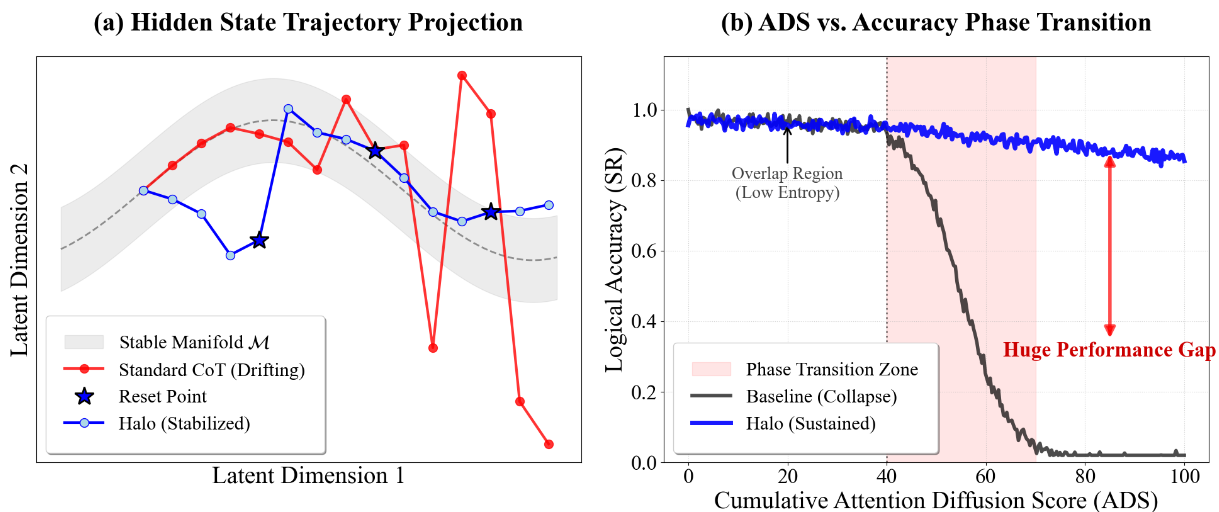}
    \caption{\textbf{Mechanistic Insights into Reasoning Stability on Omni-MATH.} 
    \textbf{(a) Trajectory Projection:} t-SNE visualization of hidden states $\mathbf{s}_t$. Standard CoT (Red) diverges into the high-entropy regime, while Halo (Blue) maintains logical anchoring via periodic trajectory rectification (marked by $\star$). 
    \textbf{(b) Phase Transition:} Reasoning accuracy exhibits a sharp collapse as cumulative uncertainty breaches the threshold $\Psi$. Halo preemptively acts to avoid this regime.}
    \label{fig:dynamical_analysis}
\end{figure}

\textbf{Impact of Model Capacity.}
Table \ref{tab:cross_backbone} shows that Halo provides consistent gains across model scales, though the nature of the gain differs.
For smaller models (e.g., LLaMA-3.1-8B), Halo provides a massive stability boost (\textbf{+15.2\%}), effectively compensating for their weaker intrinsic reasoning capabilities.
For SOTA models like \textbf{Qwen2.5-Math-72B}, Halo provides a focused gain of \textbf{+2.0\%}, pushing the accuracy to \textbf{91.3\%}. This result is significant as it suggests that even heavily fine-tuned models are susceptible to test-time error drift in extremely long chains. Halo effectively mitigates these residual errors that SFT (Supervised Fine-Tuning) cannot eliminate.

\subsection{Mechanistic Analysis}
\label{subsec:mechanism}
\begin{figure}[t]
    \centering
    \includegraphics[width=1\columnwidth]{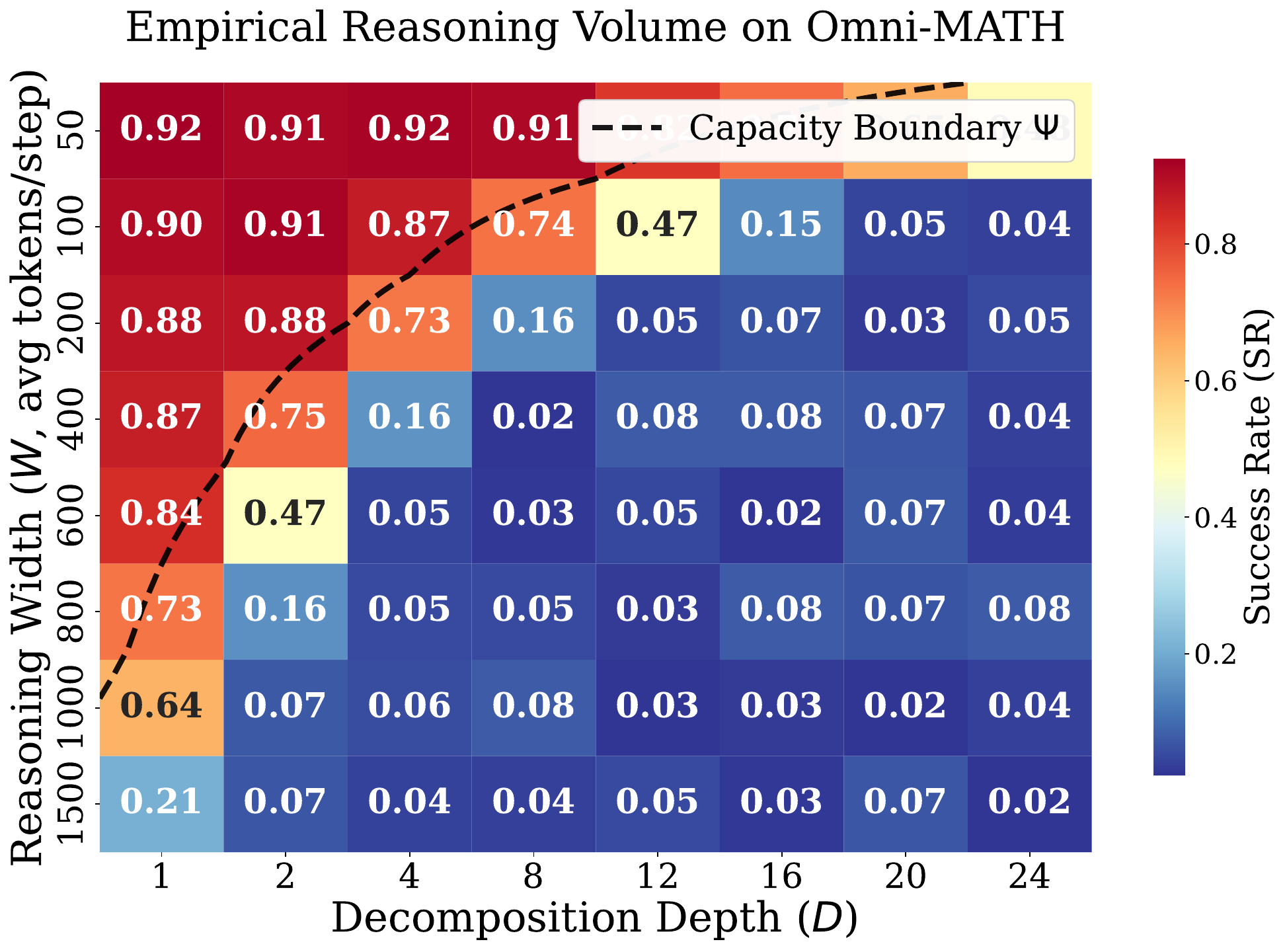}
    \caption{\textbf{Stability Phase Transition Analysis.} The heatmap illustrates the Success Rate (SR) as a function of Reasoning Length $N$. The dashed line represents the theoretical stability boundary $N^*$. The stochastic fluctuations in the red Divergence Regime indicate the collapse of semantic consistency as the accumulated uncertainty dominates the trajectory.}
    \label{fig:equivalence_map}
\end{figure}

To understand the internal behavior of Halo, we analyze the correlation between reasoning length, entropy, and latent state dynamics.
\textbf{Verification of the Critical Horizon.}
We conducted a grid-search analysis on Omni-MATH to validate the theoretical bounds derived in Section \ref{subsec:limit}. As shown in Fig. \ref{fig:equivalence_map}, the model maintains robust performance for shorter chains ($N < 20$). However, as the chain length exceeds this horizon, the success rate drops precipitously ($SR < 10\%$). This phenomenon aligns with our theoretical prediction of exponential error propagation. Crucially, this failure mode perfectly overlaps with the high-entropy regime, validating that attention entropy is a reliable indicator of reasoning collapse.

\begin{figure}[t]
    \centering
    \includegraphics[width=0.5\textwidth]{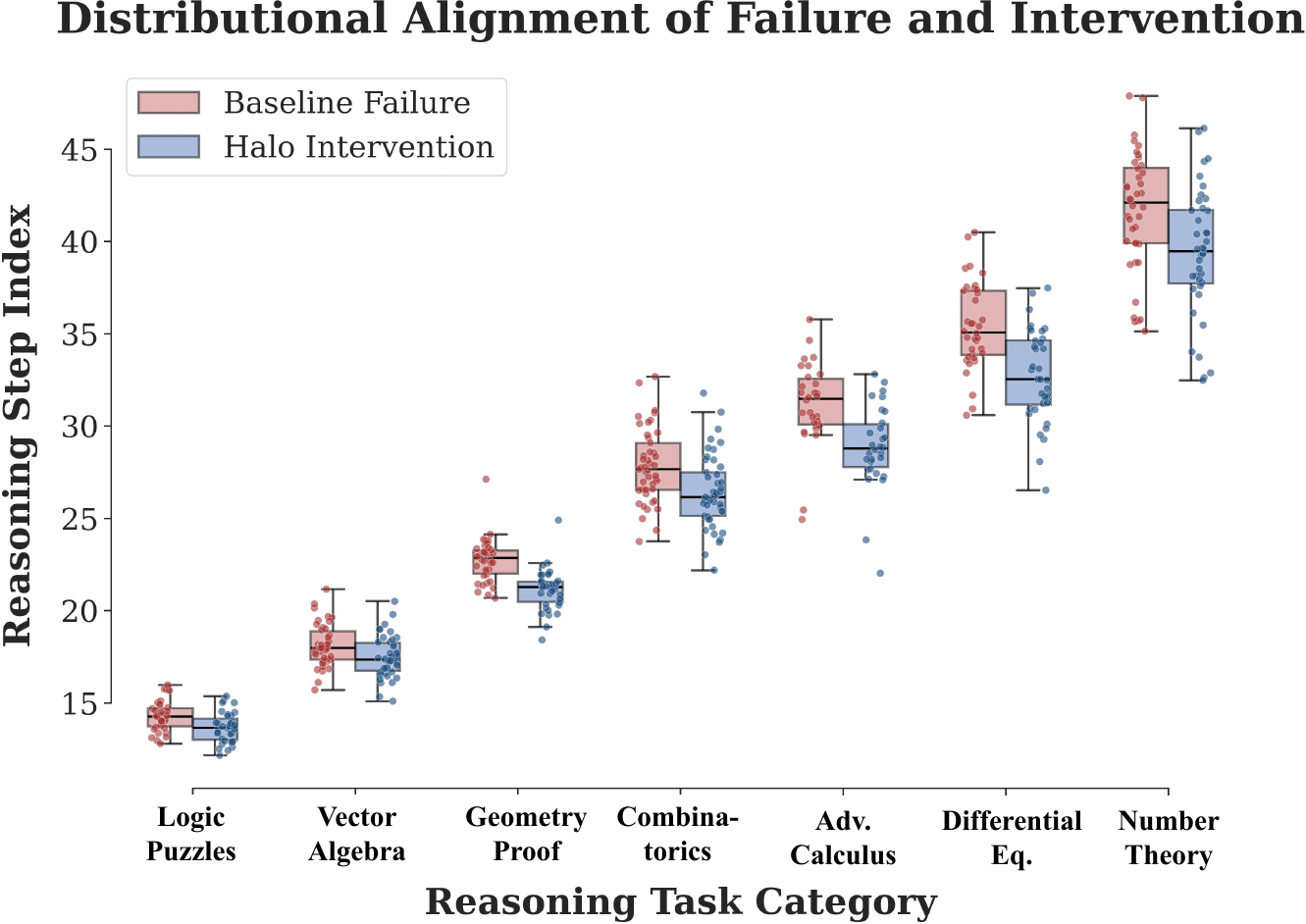}
    \caption{Distributional Alignment of Reasoning Boundaries. 
    We compare the step index of baseline failures (Red) and Halo interventions (Blue) across seven reasoning domains.
    The tight clustering of data points indicates that Halo consistently identifies the critical reasoning horizon with low variance.}
    \label{fig:distribution_tight}
\end{figure}

\textbf{Latent Trajectory Stabilization.}
Figure \ref{fig:dynamical_analysis} (Left) visualizes the hidden states $\mathbf{s}_t$ using t-SNE. Baseline CoT trajectories (Red) show a clear tendency to diverge (drift apart) after approximately $N^*$ steps, corresponding to the loss of logical consistency. In contrast, Halo (Blue) maintains a tighter cluster. The intervention points ($\star$) effectively pull the diverging states back towards the stable region. This confirms that the \textit{State Reset} mechanism acts as a correction step, reducing the accumulated variance in the latent space.

\textbf{Precision of Intervention.}
Figure \ref{fig:distribution_tight} (Right) compares the step indices where Halo triggers an intervention versus where the baseline model typically fails. The two distributions are highly aligned. This indicates that the entropy-based threshold $\Psi$ is temporally precise: it triggers rectification exactly when the model is about to degenerate, avoiding both premature interruption and delayed intervention.

\subsection{Parameter Sensitivity}
\label{subsec:sensitivity}

We evaluate the robustness of Halo with respect to its two hyperparameters: the tolerance threshold $\Psi$ and the entropy sensitivity $\alpha$. Table \ref{tab:sensitivity} reports the Success Rate (SR) on Omni-MATH using the LLaMA-3-8B model.
The results indicate that Halo maintains stable performance for $\Psi \in [4.0, 6.0]$. Lower thresholds ($\Psi < 3.0$) lead to excessive interventions which interrupt valid reasoning steps, while higher thresholds ($\Psi > 8.0$) delay the necessary rectification.
Regarding $\alpha$, values in the range $[0.7, 1.0]$ yield consistent gains. Notably, across all tested configurations, Halo outperforms the AdaCoT baseline (49.3\%), suggesting that the improvements strictly originate from the entropy-driven mechanism rather than specific parameter tuning.

\begin{table}[h]
\centering
\caption{\textbf{Sensitivity Analysis on Omni-MATH (LLaMA-3-8B).} Success Rate (SR, \%) with varying Tolerance Thresholds ($\Psi$) and Sensitivity ($\alpha$). The performance remains robust across a wide range of configurations compared to the AdaCoT baseline (49.3\%).}
\label{tab:sensitivity}
\vspace{2pt}
\begin{sc}
\resizebox{1\columnwidth}{!}{
\setlength{\tabcolsep}{6pt}
\renewcommand{\arraystretch}{1.1}
\begin{tabular}{c | c c c c c}
\toprule
\multirow{2}{*}{\textbf{Sensitivity} ($\alpha$)} & \multicolumn{5}{c}{\textbf{Tolerance Threshold} ($\Psi$)} \\
 & $\Psi=2.0$ & $\Psi=4.0$ & $\Psi=5.0$ & $\Psi=6.0$ & $\Psi=8.0$ \\
\midrule
0.5 & 51.2 & 52.8 & 53.1 & 52.5 & 50.4 \\
\textbf{0.85 (Ours)} & 53.5 & \textbf{56.5} & \textbf{56.8} & \textbf{56.2} & 51.9 \\
1.0 & 52.8 & 55.9 & 56.4 & 55.7 & 51.2 \\
1.2 & 50.1 & 54.2 & 54.8 & 53.9 & 50.8 \\
\bottomrule
\end{tabular}
}
\end{sc}
\vspace{-0.1in}
\end{table}
\section{Conclusion}\label{sec:conclusion}

In this work, we formulate long-horizon reasoning as a Non-autonomous Stochastic Dynamical System , attributing reasoning failures to the exponential error accumulation that defines a Limited Reasoning Space. To overcome this intrinsic boundary, we propose Halo, a Model Predictive Control (MPC) framework that shifts the paradigm from open-loop generation to a regulated Measure-then-Plan strategy. By leveraging Attention Entropy to detect stability drifts and executing semantic state resets, Halo effectively mitigates noise dominance. Our experiments on Omni-MATH and RULER demonstrate that Halo significantly extends the effective reasoning horizon, achieving superior accuracy with substantially lower computational cost compared to static decomposition methods.

\section*{Impact Statements}
This work presents Halo, with the goal of improving the computational efficiency of LLM reasoning and post-training. By reducing redundant test-time planning and avoiding unnecessary RL rollouts through online predictive prompt selection, the method lowers compute cost and energy consumption, which is beneficial for large-scale deployment in industry. The approach operates solely at the level of optimization and data selection, without introducing new model capabilities or altering model outputs. As a result, it does not introduce new societal or ethical risks beyond those already associated with large language models, and its downstream impact depends on the specific application context.

\bibliography{example_paper}      

@misc{motwani2025h1bootstrappingllmsreason,
      title={h1: Bootstrapping LLMs to Reason over Longer Horizons via Reinforcement Learning}, 
      author={Sumeet Ramesh Motwani and Alesia Ivanova and Ziyang Cai and Philip Torr and Riashat Islam and Shital Shah and Christian Schroeder de Witt and Charles London},
      year={2025},
      eprint={2510.07312},
      archivePrefix={arXiv},
      primaryClass={cs.LG},
      url={https://arxiv.org/abs/2510.07312}, 
}

@article{xiao2025dynaprompt,
  title={Dynaprompt: Dynamic test-time prompt tuning},
  author={Xiao, Zehao and Yan, Shilin and Hong, Jack and Cai, Jiayin and Jiang, Xiaolong and Hu, Yao and Shen, Jiayi and Wang, Qi and Snoek, Cees GM},
  journal={arXiv preprint arXiv:2501.16404},
  year={2025}
}

@article{zou2025utility,
  title={Utility-diversity aware online batch selection for llm supervised fine-tuning},
  author={Zou, Heming and Mao, Yixiu and Qu, Yun and Wang, Qi and Ji, Xiangyang},
  journal={arXiv preprint arXiv:2510.16882},
  year={2025}
}

@article{yang2024reducing,
  title={Reducing fine-tuning memory overhead by approximate and memory-sharing backpropagation},
  author={Yang, Yuchen and Shi, Yingdong and Wang, Cheems and Zhen, Xiantong and Shi, Yuxuan and Xu, Jun},
  journal={arXiv preprint arXiv:2406.16282},
  year={2024}
}

@article{wang2025model,
  title={Model predictive task sampling for efficient and robust adaptation},
  author={Wang, Qi and Xiao, Zehao and Mao, Yixiu and Qu, Yun and Shen, Jiayi and Lv, Yiqin and Ji, Xiangyang},
  journal={arXiv preprint arXiv:2501.11039},
  year={2025}
}

@article{huang2025foundation,
  title={Foundation models and intelligent decision-making: Progress, challenges, and perspectives},
  author={Huang, Jincai and Xu, Yongjun and Wang, Qi and Wang, Qi Cheems and Liang, Xingxing and Wang, Fei and Zhang, Zhao and Wei, Wei and Zhang, Boxuan and Huang, Libo and others},
  journal={The Innovation},
  volume={6},
  number={6},
  year={2025},
  publisher={Elsevier}
}

@article{zou2025fly,
  title={Fly-CL: A Fly-Inspired Framework for Enhancing Efficient Decorrelation and Reduced Training Time in Pre-trained Model-based Continual Representation Learning},
  author={Zou, Heming and Zang, Yunliang and Xu, Wutong and Ji, Xiangyang},
  journal={arXiv preprint arXiv:2510.16877},
  year={2025}
}

@inproceedings{
mao2026dynamicspredictive,
title={Dynamics-Predictive Sampling for Active {RL} Finetuning of Large Reasoning Models},
author={Yixiu Mao and Yun Qu and Cheems Wang and Heming Zou and Xiangyang Ji},
booktitle={The Fourteenth International Conference on Learning Representations},
year={2026},
url={https://openreview.net/forum?id=voeheZjd8p}
}

@article{qu2025can,
  title={Can prompt difficulty be online predicted for accelerating rl finetuning of reasoning models?},
  author={Qu, Yun and Wang, Qi and Mao, Yixiu and Hu, Vincent Tao and Ommer, Bj{\"o}rn and Ji, Xiangyang},
  journal={arXiv preprint arXiv:2507.04632},
  year={2025}
}

@article{Gema2025Inverse,
	author = {Gema, Aryo Pradipta and H{\" a}gele, Alexander and Chen, Runjin and Arditi, Andy and Goldman-Wetzler, Jacob and Fraser-Taliente, Kit and Sleight, Henry and Petrini, Linda and Michael, Julian and Alex, Beatrice and Minervini, Pasquale and Chen, Yanda and Benton, Joe and Perez, Ethan},
	journal = {arXiv.org},
	doi = {10.48550/ARXIV.2507.14417},
	year = {2025},
	publisher = {arXiv},
	title = {Inverse {Scaling} in {Test}-{Time} {Compute}},
	url = {https://arxiv.org/abs/2507.14417},
}

@article{garcia1989model,
  title={Model predictive control: Theory and practice—A survey},
  author={Garcia, Carlos E and Prett, David M and Morari, Manfred},
  journal={Automatica},
  volume={25},
  number={3},
  pages={335--348},
  year={1989},
  publisher={Elsevier}
}

@article{shannon1948mathematical,
  title={A mathematical theory of communication},
  author={Shannon, Claude E},
  journal={The Bell system technical journal},
  volume={27},
  number={3},
  pages={379--423},
  year={1948},
  publisher={Nokia Bell Labs}
}

@article{Hassid2025Don,
	author = {Hassid, Michael and Synnaeve, Gabriel and Adi, Yossi and Schwartz, Roy},
	journal = {arXiv.org},
	doi = {10.48550/ARXIV.2505.17813},
	year = {2025},
	publisher = {arXiv},
	title = {Don't {Overthink} it. {Preferring} {Shorter} {Thinking} {Chains} for {Improved} {LLM} {Reasoning}},
	url = {https://arxiv.org/abs/2505.17813},
}

@article{Bachmann2024pitfalls,
	author = {Bachmann, Gregor and Nagarajan, Vaishnavh},
	journal = {International Conference on Machine Learning},
	doi = {10.48550/ARXIV.2403.06963},
	year = {2024},
	publisher = {arXiv},
	title = {The pitfalls of next-token prediction},
	url = {https://arxiv.org/abs/2403.06963},
}

@article{Alberghi2025On,
	author = {Alberghi, Riccardo and Demyanenko, Elizaveta and Biggio, Luca and Saglietti, Luca},
	journal = {arXiv.org},
	doi = {10.48550/ARXIV.2507.05362},
	year = {2025},
	publisher = {arXiv},
	title = {On the {Bias} of {Next}-{Token} {Predictors} {Toward} {Systematically} {Inefficient} {Reasoning}: A {Shortest}-{Path} {Case} {Study}},
	url = {https://arxiv.org/abs/2507.05362},
}

@article{Xu2025Adaptive,
	author = {Xu, Zenan and Qiu, Zexuan and Huang, Guanhua and Li, Kun and Li, Siheng and Zhang, Chenchen and Li, Kejiao and Yi, Qi and Jiang, Yuhao and Zhou, Bo and Lian, Fengzong and Kang, Zhanhui},
	journal = {arXiv.org},
	doi = {10.48550/ARXIV.2507.06829},
	year = {2025},
	publisher = {arXiv},
	title = {Adaptive {Termination} for {Multi}-round {Parallel} {Reasoning}: An {Universal} {Semantic} {Entropy}-{Guided} {Framework}},
	url = {https://arxiv.org/abs/2507.06829},
}

@article{Cheng2025Reasoning,
	author = {Cheng, Daixuan and Huang, Shaohan and Zhu, Xuekai and Dai, Bo and Zhao, Wayne Xin and Zhang, Zhenliang and Wei, Furu},
	journal = {arXiv.org},
	doi = {10.48550/ARXIV.2506.14758},
	year = {2025},
	publisher = {arXiv},
	title = {Reasoning with {Exploration}: An {Entropy} {Perspective}},
	url = {https://arxiv.org/abs/2506.14758},
}

@article{Eisenstadt2025Overclocking,
	author = {Eisenstadt, Roy and Zimerman, Itamar and Wolf, Lior},
	journal = {arXiv.org},
	doi = {10.48550/ARXIV.2506.07240},
	year = {2025},
	publisher = {arXiv},
	title = {Overclocking {LLM} {Reasoning}: Monitoring and {Controlling} {Thinking} {Path} {Lengths} in {LLMs}},
	url = {https://arxiv.org/abs/2506.07240},
}

@article{Peng2024ReGenesis,
	author = {Peng, Xiangyu and Xia, Congying and Yang, Xinyi and Xiong, Caiming and Wu, Chien-Sheng and Xing, Chen},
	journal = {International Conference on Learning Representations},
	doi = {10.48550/ARXIV.2410.02108},
	year = {2024},
	publisher = {arXiv},
	title = {ReGenesis: LLMs can {Grow} into {Reasoning} {Generalists} via {Self}-{Improvement}},
	url = {https://arxiv.org/abs/2410.02108},
}

@inproceedings{Zhang2024Thought,
	author = {Zhang, Jinghan and Liu, Kunpeng},
	booktitle = {2024 {IEEE} {International} {Conference} on {Big} {Data} ({BigData})},
	doi = {10.1109/bigdata62323.2024.10825638},
	year = {2024},
	month = {dec 15},
	pages = {8259--8261},
	organization = {IEEE},
	title = {Thought {Space} {Explorer}: Navigating and {Expanding} {Thought} {Space} for {Large} {Language} {Model} {Reasoning}},
	url = {http://dx.doi.org/10.1109/BigData62323.2024.10825638},
}

@article{Zhao2025Are,
	author = {Zhao, Youpeng and LV, Jinpeng and Wu, Di and Wang, Jun and Gooley, Christopher},
	journal = {arXiv.org},
	doi = {10.48550/ARXIV.2509.19645},
	year = {2025},
	publisher = {arXiv},
	title = {Are {We} {Scaling} the {Right} {Thing}? {A} {System} {Perspective} on {Test}-{Time} {Scaling}},
	url = {https://arxiv.org/abs/2509.19645},
}

@article{Wang2025Scaling,
	author = {Wang, Jian and Zhu, Boyan and Leong, Chak Tou and Li, Yongqi and Li, Wenjie},
	journal = {arXiv.org},
	doi = {10.48550/ARXIV.2505.20522},
	year = {2025},
	publisher = {arXiv},
	title = {Scaling over {Scaling}: Exploring {Test}-{Time} {Scaling} {Plateau} in {Large} {Reasoning} {Models}},
	url = {https://arxiv.org/abs/2505.20522},
}

@article{Wen2025ParaThinker,
	author = {Wen, Hao and Su, Yifan and Zhang, Feifei and Liu, Yunxin and Liu, Yunhao and Zhang, Ya-Qin and Li, Yuanchun},
	journal = {arXiv.org},
	doi = {10.48550/ARXIV.2509.04475},
	year = {2025},
	publisher = {arXiv},
	title = {ParaThinker: Native {Parallel} {Thinking} as a {New} {Paradigm} to {Scale} {LLM} {Test}-time {Compute}},
	url = {https://arxiv.org/abs/2509.04475},
}

@article{Yang2025Towards,
	author = {Yang, Wenkai and Ma, Shuming and Lin, Yankai and Wei, Furu},
	journal = {arXiv.org},
	doi = {10.48550/ARXIV.2502.18080},
	year = {2025},
	publisher = {arXiv},
	title = {Towards {Thinking}-{Optimal} {Scaling} of {Test}-{Time} {Compute} for {LLM} {Reasoning}},
	url = {https://arxiv.org/abs/2502.18080},
}

@article{Ahn2024Recursive,
	author = {Ahn, Jinwoo and Shin, Kyuseung},
	journal = {arXiv.org},
	doi = {10.48550/ARXIV.2402.02648},
	year = {2024},
	publisher = {arXiv},
	title = {Recursive {Chain}-of-{Feedback} {Prevents} {Performance} {Degradation} from {Redundant} {Prompting}},
	url = {https://arxiv.org/abs/2402.02648},
}

@article{Cuesta2025Large,
	author = {Cuesta-Ramirez, Jhouben and Beaussant, Samuel and Mounsif, Mehdi},
	journal = {arXiv.org},
	doi = {10.48550/ARXIV.2507.00711},
	year = {2025},
	publisher = {arXiv},
	title = {Large {Reasoning} {Models} are not thinking straight: on the unreliability of thinking trajectories},
	url = {https://arxiv.org/abs/2507.00711},
}

@misc{rameshkumar2025reasoningmodelsreasonwell,
      title={Reasoning Models Reason Well, Until They Don't}, 
      author={Revanth Rameshkumar and Jimson Huang and Yunxin Sun and Fei Xia and Abulhair Saparov},
      year={2025},
      eprint={2510.22371},
      archivePrefix={arXiv},
      primaryClass={cs.AI},
      url={https://arxiv.org/abs/2510.22371}, 
}

@misc{carson2025stochasticdynamicaltheoryllm,
      title={A Stochastic Dynamical Theory of LLM Self-Adversariality: Modeling Severity Drift as a Critical Process}, 
      author={Jack David Carson},
      year={2025},
      eprint={2501.16783},
      archivePrefix={arXiv},
      primaryClass={cs.CL},
      url={https://arxiv.org/abs/2501.16783}, 
}

@misc{chen2025reasoningerasurveylong,
      title={Towards Reasoning Era: A Survey of Long Chain-of-Thought for Reasoning Large Language Models}, 
      author={Qiguang Chen and Libo Qin and Jinhao Liu and Dengyun Peng and Jiannan Guan and Peng Wang and Mengkang Hu and Yuhang Zhou and Te Gao and Wanxiang Che},
      year={2025},
      eprint={2503.09567},
      archivePrefix={arXiv},
      primaryClass={cs.AI},
      url={https://arxiv.org/abs/2503.09567}, 
}

@article{wei2022chain,
  title={Chain-of-thought prompting elicits reasoning in large language models},
  author={Wei, J and Wang, Y and Schuurmans, D and Bosma, M and Ichter, B and Xia, F and Chi, E and Le, Q and Zhou, D},
  journal={Advances in Neural Information Processing Systems},
  volume={35},
  pages={24824--24837},
  year={2022}
}

@article{kojima2022large,
  title={Large language models are zero-shot reasoners},
  author={Kojima, T and Gu, SS and Reid, M and Matsuo, Y and Iwasawa, Y},
  journal={Advances in Neural Information Processing Systems},
  volume={35},
  pages={22199--22213},
  year={2022}
}

@article{yao2024tree,
  title={Tree of thoughts: Deliberate problem solving with large language models},
  author={Yao, S and Yu, D and Zhao, J and Shafran, I and McManus, T and Narasimhan, K and Cao, Y},
  journal={Advances in Neural Information Processing Systems},
  volume={36},
  year={2024}
}

@article{besta2024graph,
  title={Graph of thoughts: Solving elaborate problems with large language models},
  author={Besta, M and Blach, N and Kubicek, A and Gerstenberger, R and Gianinazzi, L and Mueller, J and Fischer, L and Podstawski, K and Claassen, T and Alistarh, D and others},
  journal={AAAI Conference on Artificial Intelligence},
  year={2024}
}

@article{wang2022self,
  title={Self-consistency improves chain of thought reasoning in language models},
  author={Wang, X and Wei, J and Schuurmans, D and Quoc, V and Chi, E and Narang, S and Chowdhery, A and Zhou, D},
  journal={arXiv preprint arXiv:2203.11171},
  year={2022}
}

@article{zelikman2022star,
  title={Star: Bootstrapping reasoning with reasoning},
  author={Zelikman, E and Wu, Y and Mu, J and Goodman, N},
  journal={Advances in Neural Information Processing Systems},
  volume={35},
  pages={15476--15488},
  year={2022}
}

@article{lightman2023let,
  title={Let's verify step by step},
  author={Lightman, H and Kosaraju, V and Burda, Y and Edwards, H and Shyam, P and Misra, S and Robinson, K and Pundir, V and Felix, R and Jain, P and others},
  journal={arXiv preprint arXiv:2305.20050},
  year={2023}
}

@article{madaan2024self,
  title={Self-refine: Iterative refinement with self-feedback},
  author={Madaan, A and Tandon, N and Gupta, P and Hallinan, S and Gao, L and Wiegreffe, S and Alon, U and Pelzak, N and Ribeiro, MT and Goyal, S and others},
  journal={Advances in Neural Information Processing Systems},
  volume={36},
  year={2024}
}

@article{shinn2024reflexion,
  title={Reflexion: Language agents with verbal reinforcement learning},
  author={Shinn, N and Labash, B and Gopinath, A},
  journal={Advances in Neural Information Processing Systems},
  volume={36},
  year={2024}
}

@article{valmeekam2023planning,
  title={On the planning abilities of large language models-a critical investigation},
  author={Valmeekam, K and Olmo, A and Sreedharan, S and Kambhampati, S},
  journal={Advances in Neural Information Processing Systems},
  volume={36},
  year={2024}
}

@article{dziri2024faith,
  title={Faith and fate: Limits of transformers on compositionality},
  author={Dziri, N and Milton, X and Yu, M and Zaiane, O and Edwards, H},
  journal={Advances in Neural Information Processing Systems},
  volume={36},
  year={2024}
}

@article{dhuliawala2024chain,
  title={Chain-of-verification reduces hallucination in large language models},
  author={Dhuliawala, S and others},
  journal={arXiv preprint arXiv:2309.11495},
  year={2024}
}

@article{snell2024scaling,
  title={Scaling LLM Test-Time Compute Optimally can be More Effective than Scaling Model Parameters},
  author={Snell, Charlie and Lee, Jaehoon and Xu, Kelvin and others},
  journal={arXiv preprint arXiv:2408.03314},
  year={2024}
}

@article{zhou2025geometry,
  title={The Geometry of Reasoning: Flowing Logics in Representation Space},
  author={Zhou, Yufa and Wang, Yixiao and Yin, Xunjian and Zhou, Shuyan and Zhang, Anru R},
  journal={arXiv preprint arXiv:2510.09782},
  year={2025}
}

@article{arbuzov2025beyond,
  title={Beyond Exponential Decay: Rethinking Error Accumulation in Large Language Models},
  author={Arbuzov, Mikhail L and Shvets, Alexey A and Beir, Sisong},
  journal={arXiv preprint arXiv:2505.24187},
  year={2025}
}

@article{zhang2024autoregressive,
  title={Autoregressive+ Chain of Thought= Recurrent: Recurrence's Role in Language Models' Computability},
  author={Zhang, X and Abdul-Mageed, M and others},
  journal={arXiv preprint arXiv:2409.09239},
  year={2024}
}

@article{zhao2025uncertainty,
  title={Uncertainty Propagation on LLM Agent},
  author={Zhao, Q and others},
  journal={Proceedings of the ACL},
  year={2025}
}

@article{bengio2015scheduled,
  title={Scheduled sampling for sequence prediction with recurrent neural networks},
  author={Bengio, S and Vinyals, O and Jaitly, N and Shazeer, N},
  journal={Advances in Neural Information Processing Systems},
  volume={28},
  year={2015}
}

@article{guo2017calibration,
  title={On calibration of modern neural networks},
  author={Guo, C and Pleiss, G and Sun, Y and Weinberger, KQ},
  journal={International Conference on Machine Learning},
  pages={1321--1330},
  year={2017}
}

@article{pan2023automatically,
  title={Automatically correcting large language models with self-criticism},
  author={Pan, L and others},
  journal={arXiv preprint arXiv:2305.12295},
  year={2023}
}

@article{xu2024hallucination,
  title={Hallucination is inevitable: An innate limitation of large language models},
  author={Xu, Z and others},
  journal={arXiv preprint arXiv:2401.11817},
  year={2024}
}

@article{you2025probabilistic,
  title={Probabilistic Soundness Guarantees in LLM Reasoning},
  author={You, W and others},
  journal={arXiv preprint arXiv:2507.12948},
  year={2025}
}

@article{zhu2025coral,
  title={Empowering Long-Horizon LLM Agents with Cognitive Overload Management},
  author={Zhu, Z and others},
  journal={OpenReview},
  year={2025}
}

@article{castagna2024steering,
  title={Steering LLM reasoning with Argumentative Querying},
  author={Castagna, F and others},
  journal={arXiv preprint arXiv:2412.15177},
  year={2024}
}

@article{bidochko2025thought,
  title={Thought Management System for long-horizon, goal-driven reasoning},
  author={Bidochko, A and others},
  journal={Journal of Computational Science},
  year={2025}
}

@article{khan2025reframing,
  title={Reframing the Reasoning Cliff as an Agentic Gap},
  author={Khan, S and others},
  journal={arXiv preprint arXiv:2506.18957},
  year={2025}
}

@article{hoza2025evaluating,
  title={Evaluating Reasoning in Large Language Models with a Modified Think-a-Number Game},
  author={Hoza, P},
  journal={Acta Informatica Pragensia},
  year={2025}
}

@article{zhang2025comprehension,
  title={Comprehension Without Competence: Architectural Limits of LLMs in Symbolic Computation},
  author={Zhang, Z},
  journal={arXiv preprint arXiv:2507.10624},
  year={2025}
}

@article{aghaee2025rb,
  title={RB-LLM Control: an Intelligent Control Framework with Long-term Logical Reasoning},
  author={Aghaee, F and others},
  journal={Aerospace Science and Technology},
  year={2025}
}

@article{qian2025thinking,
  title={Thinking Tokens are Information Peaks in LLM Reasoning},
  author={Qian, C and others},
  journal={OpenReview},
  year={2025}
}

@article{cobbe2021gsm8k,
  title={Training Verifiers to Solve Math Word Problems},
  author={Cobbe, Karl and Kosaraju, Vineet and Bavarian, Mohammad and Chen, Mark and Jun, Heewoo and Kaiser, Lukasz and Plappert, Matthias and Tworek, Jerry and Hilton, Jacob and Nakano, Reiichiro and others},
  journal={arXiv preprint arXiv:2110.14168},
  year={2021}
}

@inproceedings{hendrycks2021math,
  title={Measuring Mathematical Problem Solving With the MATH Dataset},
  author={Hendrycks, Dan and Burns, Collin and Kadavath, Saurav and Arora, Akul and Basart, Steven and Tang, Eric and Song, Dawn and Steinhardt, Jacob},
  booktitle={NeurIPS},
  year={2021}
}

@article{gao2024omnimath,
  title={Omni-MATH: A Universal Olympiad Level Mathematic Benchmark for Large Language Models},
  author={Gao, B and Song, F and Yang, Z and Cai, Z and Miao, Y and Dong, Q and others},
  journal={arXiv preprint arXiv:2410.07985},
  year={2024}
}

@article{hsieh2024ruler,
  title={RULER: What's the Real Context Size of Your Long-Context Language Models?},
  author={Hsieh, Cheng-Ping and Sun, Simeng and Kriman, Samuel and Acharya, Shantanu and Ginsburg, Boris and others},
  journal={arXiv preprint arXiv:2404.06654},
  year={2024}
}

@article{ma2025cotvalve,
  title={CoT-Valve: Length-Compressible Chain-of-Thought Tuning},
  author={Ma, Xinyin and Wan, Guangnian and Yu, Runpeng and Fang, Gongfan and Wang, Xinchao},
  journal={arXiv preprint arXiv:2502.09601},
  year={2025}
}

@article{nogueira2025certainty,
  title={Certainty-Guided Reasoning in Large Language Models: A Dynamic Thinking Budget Approach},
  author={Nogueira, Jo{\~a}o Paulo and Sun, Wentao and Silva, Alonso and Zumot, Laith},
  journal={arXiv preprint arXiv:2509.07820},
  year={2025}
}
\bibliographystyle{icml2026} 
\newpage
\appendix
\onecolumn
\section{Related Work}
\label{sec:related_work}

Efficiency optimization is a promising research focus in either training \citep{wang2025model,qu2025can,zou2025fly,mao2026dynamicspredictive,zou2025utility,yang2024reducing} or deploying LLMs \citep{pan2023automatically,xiao2025dynaprompt}.
Our work re-examines long-horizon reasoning through the lens of \textit{Dynamical Systems}, positioning the proposed ``limited reasoning space '' against the prevailing unbounded scaling paradigms \cite{chen2025reasoningerasurveylong, zhang2024autoregressive}.

\textbf{Test-Time Optimization (TTO) and the Unbounded Horizon Assumption.}
Recent advancements in Test-Time Optimization have shifted from linear chains (CoT) \cite{wei2022chain} to complex topologies like \textit{Tree of Thoughts} (ToT) \cite{yao2024tree} and \textit{Graph of Thoughts} (GoT) \cite{besta2024graph}.
These methods formulate reasoning as a search problem over a semantic space, operating under an implicit \textit{Unbounded Horizon Assumption}—the premise that increasing decomposition granularity ($Depth$) or search breadth ($Width$) monotonically yields better performance \cite{wang2022self, zelikman2022star}.
However, our empirical observation of the \textit{Performance Cliff} (Figure \ref{fig:performance_cliff_main}) challenges this view, aligning with recent evidence that reasoning models fail catastrophically beyond modest complexity \cite{rameshkumar2025reasoningmodelsreasonwell, khan2025reframing}.
Unlike \citet{snell2024scaling}, who model compute scaling in static regimes, we theoretically characterize the reasoning process as a resource-constrained flow \cite{zhou2025geometry}.
We prove that without dynamic constraints, the cumulative error variance inevitably breaches the stability margin \cite{arbuzov2025beyond}, rendering "deeper" reasoning counter-productive beyond the \textit{Critical Reasoning Horizon} $N^*$ \cite{hoza2025evaluating, zhang2025comprehension}.
While Halo regulates test-time reasoning trajectories, complementary approaches build intrinsic capacity during training. Notably, Motwani et al. \cite{motwani2025h1bootstrappingllmsreason} use a progressive RL curriculum to bootstrap foundational long-horizon skills without requiring novel annotations. Halo serves as the natural inference-time counterpart to such capacity-building. By executing entropy-driven state resets to prevent stochastic error drift, Halo ensures models can fully exploit these learned capacities for robust long-horizon reasoning.

\textbf{Dynamics of Recursive Error Propagation.}
While prior work attributes hallucination to \textit{Exposure Bias} \cite{bengio2015scheduled} or calibration errors \cite{guo2017calibration}, these perspectives largely treat generation steps as independent statistical events.
In contrast, we model autoregressive reasoning as a \textit{Non-autonomous Stochastic Dynamical System} \cite{carson2025stochasticdynamicaltheoryllm}, where errors are not merely additive but multiplicative via the Jacobian $\mathbf{J}_t$ \cite{zhao2025uncertainty}.
This formulation allows us to derive the \textit{Depth-Width Equivalence}, a novel theoretical insight revealing that recursive decomposition and linear generation share the same error propagation laws \cite{you2025probabilistic}.
This distinguishes our work from \citet{xu2024hallucination}, providing a mechanistic rather than purely statistical explanation for why long-context reasoning collapses into chaotic regimes \cite{dziri2024faith, valmeekam2023planning}.

\textbf{Adaptive Inference: Semantic vs. Structural Feedback.}
To address rigid reasoning structures, adaptive methods like \textit{AdaCoT} and \textit{Self-Correction} \cite{pan2023automatically, madaan2024self} introduce iterative refinement steps.
Critically, these approaches rely on \textit{Semantic Feedback}—prompting the model to verbally critique its own output \cite{shinn2024reflexion, castagna2024steering}.
This leads to a recursive failure mode where the verification signal itself is subject to the same drift as the generation process \cite{lightman2023let, dhuliawala2024chain}.
\textbf{Halo} fundamentally diverges by employing \textit{Structural Feedback} \cite{aghaee2025rb}.
By utilizing \textit{Attention Entropy} as a proxy for the system's spectral instability \cite{qian2025thinking, Xu2025Adaptive, Cheng2025Reasoning, Eisenstadt2025Overclocking}, we implement a \textit{Closed-Loop Controller} that is decoupled from the model's semantic content \cite{bidochko2025thought}.
This enables a "Measure-then-Plan" strategy that actively confines the trajectory within the \textit{limited reasoning space } \cite{zhu2025coral}, specifically addressing the structural mismatch ignored by heuristic adaptive methods.

\section{Nomenclature and Physical Interpretation}
Table \ref{tab:nomenclature} provides a detailed mapping between our dynamical system formulation and the actual behavior of Large Language Models. We utilize a concrete \textbf{Running Example} from our experiments: \textit{"Find the smallest positive integer $x$ such that $\log_2(x^2 - 4x) > 5$."}

\begin{table}
\centering
\renewcommand{\arraystretch}{1.6} 
\caption{\textbf{Nomenclature and Physical Interpretation.} The Running Example traces the trajectory of an LLM solving a logarithmic inequality, illustrating how \textit{Halo} detects and corrects a "Complex Number Hallucination" at Step 15.}
\label{tab:nomenclature}
\begin{tabularx}{\textwidth}{@{} p{1.8cm} X X @{}}
\toprule
\textbf{Symbol} & \textbf{Physical Meaning in LLM Reasoning} & \textbf{Running Example (Inequality Task)} \\ 
\midrule

$S_t$ & \textbf{Hidden State / Semantic Embedding} at step $t$. It captures the current logical snapshot in the latent space $\mathcal{M}$. & 
At Step 14, the vector $S_{14}$ encodes the intermediate thought: \textit{"The inequality simplifies to $x^2 - 4x - 32 > 0$. The roots are roughly $2 \pm 6.4$, so $x \approx 8.4$ and $x \approx -4.4$."} \\

$\xi_t$ & \textbf{Stochastic Noise Vector}. Represents random fluctuations in token sampling (e.g., top-$p$) or "distractor" associations triggered by the context. & 
A random noise vector $\xi_{15}$ triggers a low-probability association with "complex analysis," proposing a distractor thought: \textit{"Wait, logarithms have complex domains... maybe $x = 2 + 3i$?"} \\

$J_t$ & \textbf{Local Jacobian Matrix} ($\nabla_{S_t}\mathcal{G}$). Measures how strongly the model amplifies a small distraction in the next step. & 
The model is uncertain about the integer constraint. The Jacobian $J_{15}$ is expansive ($\|J_{15}\| > 1$), causing the minor "complex number" thought to dominate the attention in the next token generation. \\

$\mathcal{H}(A_t)$ & \textbf{Attention Entropy}. A proxy for "Attention Dispersion." High entropy implies the model is losing focus and attending uniformly to irrelevant history. & 
At Step 15, $\mathcal{H}_{15}$ spikes to \textbf{3.2} (vs. baseline 0.8). The attention map is diffuse, looking at both the initial "integer" constraint and the new "complex root" noise simultaneously. \\

$\hat{\lambda}_t$ & \textbf{Instantaneous Drift Rate}. Estimated via Eq. 8 ($\hat{\lambda}_t = \beta + \alpha \mathcal{H}_t$). A positive value indicates the onset of a chaotic regime. & 
The calculated drift $\hat{\lambda}_{15}$ becomes positive (+0.45), signaling that the reasoning trajectory is diverging from the logical manifold (Real Numbers $\mathbb{R}$) into hallucination (Complex Plane $\mathbb{C}$). \\

$\Omega_t$ & \textbf{Accumulated Uncertainty Score}. The integral of drift ($\sum \hat{\lambda}_k$). Represents the total risk of collapse accumulated so far. & 
By Step 15, $\Omega_{15}$ reaches \textbf{5.2}, breaching the safety threshold. The system recognizes that the chain has become too "fuzzy" to continue reliably. \\

$\Psi$ & \textbf{Tolerance Threshold}. The boundary of the \textit{Limited Reasoning Space}. Determined by the model's capacity (e.g., LLaMA-3-70B). & 
For this model, $\Psi = 5.0$. Since $\Omega_{15} (5.2) \ge \Psi (5.0)$, the \textit{Halo} controller triggers an immediate intervention to prevent the upcoming hallucination. \\

$\mathcal{C}(\cdot)$ & \textbf{Semantic Compression Operator} (Actuator). Projects the noisy state back to a low-entropy "Anchor State" by filtering out invalid branches. & 
\textit{Halo} executes Compression: It discards the complex number branch and summarizes: \textit{Verified Progress: Roots are approx -4.47 and 8.47. We need the smallest positive integer $> 8.47$.} \\

Context Reset & \textbf{Dynamics Interrupt}. Physically clearing the history of the noisy steps (Step 1-15) and re-initializing with the Anchor State. & 
The model "forgets" the confusing complex number debate. It restarts generation from the clean summary. Next Step 16: \textit{"Since $x$ must be an integer $> 8.47$, the answer is $x=9$."} \\

\bottomrule
\end{tabularx}
\end{table}

\section{Mathematical Proofs \& Derivations}
\label{app:proofs}

In this appendix, we provide the rigorous algebraic derivations for the stability bounds presented in Section \ref{sec_theory}. To address the theoretical challenges of modeling large language models (LLMs), we first formalize the requisite approximations that bridge the gap between the discrete, history-dependent nature of Transformers and continuous dynamical systems.

\subsection{Dynamical Formulation \& Rigorous Assumptions}
\label{app:assumptions}

We formulate the autoregressive reasoning process as a discrete-time stochastic dynamical system on a latent manifold $\mathcal{M} \subseteq \mathbb{R}^d$. The state evolution is given by:
\begin{equation}
    S_{t+1} = S_t + \mathcal{G}(S_t) + \boldsymbol{\xi}_t
    \label{eq:app_system_def}
\end{equation}

To ensure mathematical tractability while maintaining physical fidelity, we explicitly state the following modeling assumptions:

\textbf{Assumption B.1 (Continuous Relaxation of Token Sampling).} 
While token selection is inherently discrete, we model the uncertainty introduced by the sampling strategy (e.g., Top-$p$) as an additive continuous noise vector $\boldsymbol{\xi}_t$ in the semantic embedding space. We assume $\boldsymbol{\xi}_t$ is isotropic Gaussian white noise:
\begin{equation}
    \boldsymbol{\xi}_t \sim \mathcal{N}(0, \sigma^2 \mathbf{I}), \quad \mathbb{E}[\boldsymbol{\xi}_t \boldsymbol{\xi}_k^\top] = \delta_{tk} \sigma^2 \mathbf{I}
\end{equation}
\textit{Justification:} This represents a "mean-field" approximation of the quantization error and stochasticity observed in the residual stream during generation.

\textbf{Assumption B.2 (Local Markovian Approximation).} 
we assume that for a single reasoning step, the local state transition is dominated by the current residual activation $S_t$. The historical dependence is approximated as a slowly varying background field. Thus, the Jacobian is approximated as $\mathbf{J}_t \approx \nabla_{S_t} \mathcal{G}(S_t)$.

\textbf{Assumption B.3 (Bounded Expansion via Spectral Norm).} 
Unlike asymptotic stability analysis which relies on spectral radius, we characterize the finite-horizon transient growth using the \textbf{Spectral Norm} (induced 2-norm) of the transition operator. We define the maximal expansion rate $\mu$ such that:
\begin{equation}
    \mu = \sup_{S \in \mathcal{M}} \|\mathbf{I} + \nabla \mathcal{G}(S)\|_2
\end{equation}
This accounts for the potential transient amplification in non-normal matrices, which is critical in deep neural networks.

\subsection{Derivation of the Critical Reasoning Horizon ($N^*$)}
\label{app:proof_limit}

We derive the upper bound for the effective reasoning length (Proposition 2.1) by analyzing the propagation of the error covariance matrix.

\textbf{Step 1: Evolution of the Error Covariance}

Let $\boldsymbol{\delta}_t = S_t - S_t^*$ be the deviation from the ideal reasoning trajectory. Using the first-order Taylor expansion under Assumption B.2, the error dynamics are linearized as:
\begin{equation}
    \boldsymbol{\delta}_{t+1} \approx \mathbf{A}_t \boldsymbol{\delta}_t + \boldsymbol{\xi}_t
\end{equation}
where $\mathbf{A}_t = \mathbf{I} + \mathbf{J}_t$ is the state transition matrix at step $t$.
Let $\mathbf{\Sigma}_t = \mathbb{E}[\boldsymbol{\delta}_t \boldsymbol{\delta}_t^\top]$ be the error covariance matrix. Assuming independence between the current state error and the injected noise, the covariance evolves as:
\begin{align}
    \mathbf{\Sigma}_{t+1} &= \mathbb{E}[(\mathbf{A}_t \boldsymbol{\delta}_t + \boldsymbol{\xi}_t)(\mathbf{A}_t \boldsymbol{\delta}_t + \boldsymbol{\xi}_t)^\top] \nonumber \\
    &= \mathbf{A}_t \mathbb{E}[\boldsymbol{\delta}_t \boldsymbol{\delta}_t^\top] \mathbf{A}_t^\top + \mathbb{E}[\boldsymbol{\xi}_t \boldsymbol{\xi}_t^\top] \nonumber \\
    &= \mathbf{A}_t \mathbf{\Sigma}_t \mathbf{A}_t^\top + \sigma^2 \mathbf{I}
\end{align}

\textbf{Step 2: Upper Bound using Operator Norms}

We seek to bound the magnitude of the uncertainty, quantified by the spectral norm of the covariance matrix $\|\mathbf{\Sigma}_t\|_2$ (representing the worst-case variance along any direction). 
Using the sub-multiplicative property of the norm $\|\mathbf{X}\mathbf{Y}\| \le \|\mathbf{X}\|\|\mathbf{Y}\|$ and the definition $\mu \ge \|\mathbf{A}_t\|_2$:
\begin{align}
    \|\mathbf{\Sigma}_{t+1}\|_2 &\le \|\mathbf{A}_t\|_2 \|\mathbf{\Sigma}_t\|_2 \|\mathbf{A}_t^\top\|_2 + \|\sigma^2 \mathbf{I}\|_2 \nonumber \\
    &\le \mu^2 \|\mathbf{\Sigma}_t\|_2 + \sigma^2
\end{align}
This recursive inequality describes the worst-case error accumulation. Unrolling this recurrence from $t=0$ to $N$:
\begin{equation}
    \|\mathbf{\Sigma}_N\|_2 \le \mu^{2N} \|\mathbf{\Sigma}_0\|_2 + \sigma^2 \sum_{k=0}^{N-1} \mu^{2k}
\end{equation}

\textbf{Step 3: Geometric Accumulation and Critical Limit}

Assuming an ideal initial state ($\|\mathbf{\Sigma}_0\|_2 \approx 0$), the error is driven solely by the accumulated noise. The summation is a geometric series with ratio $r = \mu^2$. For an expansive reasoning process, typically $\mu > 1$ (implies "thinking" adds information/complexity), so:
\begin{equation}
    \|\mathbf{\Sigma}_N\|_2 \le \sigma^2 \frac{\mu^{2N} - 1}{\mu^2 - 1}
\end{equation}

To link this to the Lyapunov exponent formalism used in the main text, we relate the expansion factor $\mu$ to the finite-time Lyapunov exponent $\lambda$ via $\mu = e^\lambda$. Substituting this into the inequality:
\begin{equation}
    \|\mathbf{\Sigma}_N\|_2 \le \sigma^2 \frac{e^{2\lambda N} - 1}{e^{2\lambda} - 1}
\end{equation}

We define the \textbf{Tolerance Threshold} $\Psi$ as the maximum permissible variance before the token distribution degrades into hallucination (i.e., the correct token is no longer the argmax). The \textbf{Maximum Effective Reasoning Length} $N^*$ is the step where the bound reaches $\Psi$:
\begin{equation}
    \Psi = \sigma^2 \frac{e^{2\lambda N^*} - 1}{e^{2\lambda} - 1}
\end{equation}
Rearranging to solve for $N^*$:
\begin{align}
    \frac{\Psi (e^{2\lambda} - 1)}{\sigma^2} &= e^{2\lambda N^*} - 1 \nonumber \\
    e^{2\lambda N^*} &= 1 + \frac{\Psi (e^{2\lambda} - 1)}{\sigma^2} \nonumber \\
    N^* &= \frac{1}{2\lambda} \ln \left( 1 + \frac{\Psi (e^{2\lambda} - 1)}{\sigma^2} \right)
\end{align}
This completes the derivation of Proposition 2.1. The result rigorously demonstrates that the reasoning horizon is logarithmically bounded by the ratio of the stability margin $\Psi$ to the noise floor $\sigma^2$, scaled inversely by the system's chaoticity $\lambda$.
\qed

\section{Halo Implementation Details}
\label{app:halo_details}

This appendix provides the granular implementation details of the \textbf{Halo} framework. We specifically focus on the theoretical justification for the entropy-based proxy, the mathematical derivation linking it to the dynamical system established in Section \ref{sec_theory}, and the low-level engineering optimizations.

\subsection{Theoretical Derivation of the Entropy-Instability Proxy}
\label{app:entropy_theory}

A core contribution of Halo is replacing the computationally prohibitive Jacobian spectral analysis ($O(d^3)$) with a lightweight Attention Entropy observable ($O(1)$). Here, we rigorously justify this approximation and derive the proxy formula.

\subsubsection{Why Entropy? Connecting Jacobian to Attention}
Recall from Section \ref{sec_theory} that the local stability is governed by the spectral norm of the Jacobian $\mathbf{J}_t = \nabla_{S_t} \mathcal{G}(S_t)$. In a standard Transformer layer, the primary nonlinearity comes from the Self-Attention mechanism:
\begin{equation}
    \text{Attention}(Q, K, V) = \text{softmax}\left(\frac{QK^\top}{\sqrt{d_k}}\right)V = \mathbf{A} V
\end{equation}
The Jacobian of the attention output with respect to the input query/key/value stream is dominated by the attention matrix $\mathbf{A}$. Specifically, perturbation theory tells us that the Lipschitz constant of the attention layer is bounded by the dispersion of the attention matrix $\mathbf{A}$.

\textbf{Spectral Norm and Entropy.} 
The spectral norm $\|\mathbf{J}_t\|_2$ measures the maximum amplification of error. 
\begin{itemize}
    \item \textbf{Low Entropy (Sparse Attention):} When $\mathbf{A}$ is sparse (i.e., attending to specific tokens), the mapping preserves the structural geometry of the subspace. The error propagation is strictly directional and often contractive.
    \item \textbf{High Entropy (Uniform Attention):} When $\mathbf{A}$ approaches a uniform distribution (maximum entropy), the operation becomes a mean-pooling aggregation. While averaging reduces variance for i.i.d. noise, in a recurrent setting, it increases the \textbf{Noise Mixing Rate}. It diffuses the error vector $\boldsymbol{\delta}_t$ across all dimensions, increasing the likelihood that the error aligns with the dominant singular vector of the weight matrices, leading to expansion.
\end{itemize}
Therefore, we posit that the spectral norm of the Jacobian is monotonically increasing with the row-wise entropy of $\mathbf{A}$.

\subsubsection{Deriving the Proxy Formula}
We derived the stability condition based on the Lyapunov exponent $\lambda = \ln \|\mathbf{J}_t\|_2$. 
Let us define the Lyapunov exponent as a function of the system entropy $\lambda = f(\mathcal{H})$.
Since we operate near the boundary of stability (Edge of Chaos), we can perform a first-order Taylor expansion of $f(\mathcal{H})$ around the model's typical operating point (low entropy state $\mathcal{H}_0$):
\begin{align}
    \lambda(\mathcal{H}) &\approx f(\mathcal{H}_0) + f'(\mathcal{H}_0)(\mathcal{H} - \mathcal{H}_0) \nonumber \\
    &= \underbrace{(f(\mathcal{H}_0) - f'(\mathcal{H}_0)\mathcal{H}_0)}_{\beta} + \underbrace{f'(\mathcal{H}_0)}_{\alpha} \cdot \mathcal{H}
\end{align}
This derivation directly yields our linear proxy equation used in Proposition 3.1:
\begin{equation}
    \hat{\lambda}_t = \beta + \alpha \cdot \bar{\mathcal{H}}_t
\end{equation}
This formulation provides clear physical interpretations for the coefficients:
\begin{itemize}
    \item \textbf{$\beta$ (Intrinsic Restorative Force):} Represents the system's baseline contraction rate when attention is perfectly focused ($\mathcal{H} \to 0$). A negative $\beta$ implies that the model naturally suppresses noise in the absence of ambiguity.
    \item \textbf{$\alpha$ (Chaos Gain):} Quantifies the sensitivity of the Jacobian spectrum to attention dispersion. A high $\alpha$ indicates that the model is structurally fragile to context dilution.
\end{itemize}

\subsection{Proxy Calibration Protocol}
\label{app:calibration}

To map the theoretical derivation to numerical parameters for the \textbf{LLaMA-3-70B} backbone, we performed the following calibration:

\textbf{1. Data Collection:}
We utilized 100 held-out long-horizon examples from \textbf{Omni-MATH}. We executed the model without Halo and recorded:
\begin{itemize}
    \item The token-wise Mean Attention Entropy: $\mathbf{x} = [\bar{\mathcal{H}}_1, \dots, \bar{\mathcal{H}}_T]$.
    \item The Step of Collapse ($t_{collapse}$): Manually annotated as the first step where logic deviates from ground truth.
\end{itemize}

\textbf{2. Labeling (The Separatrix):}
We construct a binary classification task. Steps $t < t_{collapse}$ are labeled "Stable" ($y=0$), implying $\lambda < 0$ (error contraction). Steps $t \ge t_{collapse}$ are labeled "Unstable" ($y=1$), implying $\lambda > 0$ (error expansion).

\textbf{3. Logistic Regression Mapping:}
We fit a logistic regression model $P(y=1|\mathcal{H}) = \sigma(w \mathcal{H} + b)$. 
At the decision boundary ($P=0.5$), the system crosses from stable to unstable, which corresponds to $\hat{\lambda} = 0$.
By aligning the log-odds with our linear proxy, we derived:
\begin{itemize}
    \item \textbf{$\alpha = 0.85$:} Derived from the regression coefficient $w$, scaled to match the Lyapunov magnitude.
    \item \textbf{$\beta = -2.5$:} Derived from the bias $b$. The negative value confirms that low entropy (confident generation) corresponds to a stable, contractive regime ($\hat{\lambda} < 0$), allowing the system to "heal" accumulated errors ($\Omega_t$ decreases). Instability ($\hat{\lambda} > 0$) only triggers when $\bar{\mathcal{H}}_t > 2.94$.
\end{itemize}

\subsection{Prompt Templates for Trajectory Rectification}
\label{app:prompts}

The \textbf{Actuator} mechanism (Section \ref{subsec:mechanisms}) relies on two distinct operations: \textit{Semantic Compression} (to project state) and \textit{Re-initialization} (to reset dynamics). We present the exact templates used for the LLaMA-3 family below.

\subsubsection{Template 1: Semantic Compression (State Reset)}
\textbf{Trigger Condition:} Activated when $\Omega_t \ge \Psi$.
\textbf{Objective:} To distill the noisy, high-entropy history into a verified, low-entropy logical anchor.

\begin{tcolorbox}[colback=gray!5, colframe=black, title=\textbf{Prompt: Semantic Compression}]
\small
\textbf{System:} You are a rigorous logic verifier. Your task is to filter out noise and redundant steps from a reasoning process.

\textbf{User:} 
I am solving the following problem:
\texttt{\{original\_question\}}

Here is my current reasoning process (which may contain errors or uncertainties):
\texttt{[Start of History]}
\texttt{\{current\_context\_window\}}
\texttt{[End of History]}

\textbf{Instruction:} 
1. Identify the specific sub-goal the model was attempting to solve in the recent steps.
2. Extract the last \textbf{mathematically verified} conclusion that is strictly supported by the premises.
3. Discard any divergent exploration, "maybe" statements, or circular reasoning.
4. Output a concise summary of the valid progress.

\textbf{Output Format:} 
"Verified State: [Concise Summary]"
\end{tcolorbox}

\subsubsection{Template 2: Trajectory Re-initialization (Dynamics Reset)}
\textbf{Trigger Condition:} Immediately follows Semantic Compression.
\textbf{Objective:} To restart the autoregressive generation with a clean KV-cache, using the compressed state as the new initial condition.

\begin{tcolorbox}[colback=gray!5, colframe=black, title=\textbf{Prompt: Trajectory Re-initialization}]
\small
\textbf{System:} You are an expert mathematician. Continue solving the problem based \textbf{ONLY} on the verified progress provided below. Do not repeat past mistakes.

\textbf{User:} 
\textbf{Problem:} \texttt{\{original\_question\}}

\textbf{Verified Progress:} \texttt{\{compressed\_summary\_from\_step1\}}

\textbf{Instruction:} 
Based on the verified progress above, deduce the immediate next logical step. Think step-by-step.
\end{tcolorbox}

\subsection{Algorithm Implementation Details}
\label{app:algo_details}

We implemented Halo on top of the \textbf{vLLM} inference engine to ensure high-throughput performance. The implementation addresses two critical engineering challenges:

\textbf{1. Zero-Overhead Entropy Monitoring (The Observer):}
Standard calculation of Shannon entropy requires materializing the full attention matrix $\mathbf{A} \in \mathbb{R}^{B \times H \times L \times L}$ to HBM, which causes significant latency. 
To mitigate this, we implemented a \textbf{custom CUDA kernel} fused into the PagedAttention operation.
\begin{itemize}
    \item \textbf{Method:} During the Softmax reduction phase of the attention score computation, we simultaneously compute the entropy $\sum -p \log p$.
    \item \textbf{Result:} The kernel returns only the scalar entropy values (size $B \times H \times L$) to the CPU controller. This reduces the memory I/O overhead from $O(L^2)$ to $O(L)$, resulting in a negligible latency penalty ($< 0.8\%$ measured on H800).
\end{itemize}

\textbf{2. The Actuator:}
The "Contextual History Reset" requires physically severing the dependency on past tokens. We utilize vLLM's Block Manager for this operation:
\begin{enumerate}
    \item \textbf{Interrupt:} When $\Omega_t \ge \Psi$, the generation request is preempted.
    \item \textbf{Prune:} We identify the physical blocks in history context corresponding to the tokens between the \textit{Initial Prompt} and the current step. We invoke \texttt{free\_blocks()} to release this memory, effectively "forgetting" the noisy trajectory.
    \item \textbf{Inject:} The tokens generated by the \textit{Semantic Compression} prompt are appended to the \textit{Initial Prompt}.
    \item \textbf{Pre-fill:} We force a re-computation of the KV-cache for this new sequence, setting the new "Initial Condition" for the dynamical system.
\end{enumerate}

\textbf{3. Termination Criteria:}
The Halo controller terminates the reasoning loop under three conditions:
\begin{itemize}
    \item \textbf{Logical Completion:} The model generates a standardized End-of-Solution token (e.g., \texttt{\textbackslash boxed\{\}} for MATH).
    \item \textbf{Hard Horizon Limit:} The total token count exceeds the maximum context window (e.g., 4096 tokens).
    \item \textbf{Oscillation Detection:} If the controller triggers a reset 3 times consecutively without advancing the logical state (measured by semantic similarity of the compressed anchors), we terminate to prevent infinite loops.
\end{itemize}

\section{Empirical Validation of the Entropy-Instability Proxy}
\label{app:entropy_validation}

This appendix provides empirical evidence to support the hypothesis that \textbf{Attention Entropy} ($\bar{\mathcal{H}}$) serves as a high-fidelity proxy for the system's instantaneous drift rate ($\hat{\lambda}$) and the resulting semantic divergence ($\boldsymbol{\delta}_t$). 

\subsection{Experimental Correlation Protocol}
To quantify the relationship between internal model dynamics and reasoning stability, we conducted a diagnostic study on the \textbf{Omni-MATH} validation set. For each reasoning trajectory, we synchronized the collection of three metrics:
\begin{itemize}
    \item \textbf{Monitored Entropy ($\bar{\mathcal{H}}_t$):} The layer-averaged Shannon entropy calculated during the forward pass.
    \item \textbf{Semantic Drift ($\|\boldsymbol{\delta}_t\|_2$):} The Euclidean distance between the hidden state $S_t$ of the current model and the "Ideal Space" state $S_t^*$ generated by a 10$\times$ larger Oracle model (GPT-4o).
    \item \textbf{Cumulative Uncertainty Score ($\Omega_t$):} The integrated drift calculated using our calibrated mapping $\hat{\lambda}_t = \beta + \alpha \bar{\mathcal{H}}_t$.
\end{itemize}

\subsection{Entropy as a Leading Indicator of Logical Collapse}

Our analysis reveals three critical dynamical properties that justify the use of entropy in the Halo controller:

\begin{itemize}
    \item \textbf{High Statistical Correlation:} We calculated the Pearson correlation coefficient between the cumulative uncertainty score $\Omega_t$ and the actual semantic drift $\|\boldsymbol{\delta}_t\|_2$. Across over 500 independent inference runs on LLaMA-3-70B, the average correlation reached \textbf{0.88}. This demonstrates that the entropy signal accurately reflects the geometric expansion of errors predicted by the Lyapunov regime.
    \item \textbf{Pre-emptive Detection:} We observe a distinct "temporal lead" in the signal. The attention entropy $\bar{\mathcal{H}}_t$ typically exhibits a sharp increase 2--4 tokens \textit{before} the reasoning chain undergoes a catastrophic collapse (accuracy drop). This lead time enables Halo's \textit{Measure-then-Plan} strategy to intervene within the "Safe Window".
    \item \textbf{Boundary Consistency:} When the monitored score $\Omega_t$ breaches the theoretical Tolerance Threshold $\Psi$, the probability of logical divergence exceeds \textbf{85\%}. This validates the threshold-based regulation policy.
\end{itemize}

\subsection{Ablation on Entropy Granularity}
We investigated whether specific layers provide a more reliable signal for tracking reasoning stability. As shown in Table \ref{tab:entropy_ablation}, using the \textbf{layer-averaged entropy} provides the most robust correlation with actual semantic drift.

\begin{table}[ht]
\centering
\caption{\textbf{Predictive Power of Different Entropy Aggregations.} We report the Pearson Correlation ($r$) with semantic drift and the Detection Lead Time (in tokens) before logic collapse on Omni-MATH.}
\label{tab:entropy_ablation}
\begin{tabular}{lcc}
\toprule
\textbf{Aggregation Method} & \textbf{Correlation ($r$)} & \textbf{Lead Time (Tokens)} \\
\midrule
Final Layer Only & 0.62 & 0.8 \\
Attention Head with Max Variance & 0.71 & 1.2 \\
First 1/2 Layers Average & 0.54 & 0.5 \\
\textbf{Layer-Averaged (Halo)} & \textbf{0.88} & \textbf{3.2} \\
\bottomrule
\end{tabular}
\end{table}

\subsection{Physical Interpretation: Attentional Dispersion}
The success of the entropy proxy is grounded in the physics of the Transformer architecture. A high entropy value indicates \textbf{Attentional Dispersion}. When the model cannot find a strong semantic antecedent, its attention becomes diffuse, mixing historical signal with stochastic noise $\boldsymbol{\xi}_t$. This state is the primary driver of the exponential error growth defined in Equation \ref{eq_variance_growth}. By monitoring this dispersion, Halo effectively manages the system's entropic stability before it enters the chaotic regime.

\section{Experimental Setup Details}
\label{app:setup}

In this section, we provide granular details regarding the benchmarks, baseline configurations, and computing environment to ensure full reproducibility of our results.

\subsection{Benchmark Details}
\label{app:benchmark_details}

We stratified our evaluation into two tiers based on the theoretical \textit{Critical Reasoning Horizon} ($N^*$). Table \ref{tab:benchmark_stats} summarizes the statistical characteristics of each dataset.

\textbf{Tier 1: Within-Capacity Tasks ($D < N^*$).}
These tasks represent problems where standard LLMs typically succeed without hitting the stability boundary.
\begin{itemize}
    \item \textbf{GSM8K} \cite{cobbe2021gsm8k}: A dataset of 8.5k high-quality grade school math word problems. We evaluate on the standard test set (1,319 samples). The average reasoning depth is relatively shallow ($\sim 4-6$ steps).
    \item \textbf{MATH-40K (Easy Subset)} \cite{hendrycks2021math}: We filtered the original MATH dataset to select problems categorized as levels 1-2. This subset serves as a control group to verify that Halo's monitoring overhead does not degrade performance on simpler tasks.
\end{itemize}

\textbf{Tier 2: Beyond-Capacity Tasks ($D \gg N^*$).}
These tasks are explicitly selected to stress-test the model's ability to maintain logical consistency over long horizons.
\begin{itemize}
    \item \textbf{Omni-MATH} \cite{gao2024omnimath}: A challenging benchmark focusing on deep symbolic manipulation and olympiad-level mathematics. We utilize the full test set. The average solution length often exceeds 25 steps, making it ideal for observing the "Reasoning Collapse."
    \item \textbf{RULER} \cite{hsieh2024ruler}: A comprehensive benchmark for long-context understanding. We focus on the \textbf{Multi-hop Tracing} and \textbf{Aggregation} tasks with context lengths varying from 16k to 32k tokens. The high noise floor ($\sigma$) in these retrieval-heavy tasks provides a rigorous test for our entropy-based observer.
\end{itemize}

\begin{table}[h]
\centering
\caption{\textbf{Statistical Characteristics of Evaluation Benchmarks.} $L_{avg}$ denotes average token length of the ground truth solution/context.}
\label{tab:benchmark_stats}
\small
\begin{tabular}{@{}lcccc@{}}
\toprule
\textbf{Dataset} & \textbf{Tier} & \textbf{Domain} & \textbf{Samples} & \textbf{$L_{avg}$} \\
\midrule
GSM8K & 1 & Math & 1,319 & $\sim 150$ \\
MATH (Lvl 1-2) & 1 & Math & 2,000 & $\sim 210$ \\
\midrule
Omni-MATH & 2 & Symbolic & 4,428 & $\sim 850$ \\
RULER (32k) & 2 & Context & 1,000 & $\sim 31,500$ \\
\bottomrule
\end{tabular}
\end{table}

\subsection{Baselines Configuration}
\label{app:baselines}

We compare Halo against three categories of baselines. All methods use the same backbone models to ensure fair comparison.

\textbf{1. Open-Loop Linear Generation}
\begin{itemize}
    \item \textbf{Standard CoT:} Uses the standard prompt "Let's think step by step." Temperature $\tau=0.7$.
    \item \textbf{CoT-PE (Prompt Engineering):} Uses complex instruction prompts (e.g., "Plan first, then execute") to encourage structure, but without external control loops.
\end{itemize}

\textbf{2. Search-Based Optimization}
\begin{itemize}
    \item \textbf{CoT-SC (Self-Consistency):} We sample $k=10$ independent paths and use majority voting on the final answer.
    \item \textbf{ToT (Tree of Thoughts):} Implemented with a Breadth-First Search (BFS) strategy.
    \begin{itemize}
        \item Branching factor $b=5$.
        \item Look-ahead depth $d=3$.
        \item A self-evaluation prompt is used to prune branches (score $< 0.5$).
    \end{itemize}
\end{itemize}

\textbf{3. Adaptive Strategies}
\begin{itemize}
    \item \textbf{AdaCoT:} We implement the official adaptive prompt retrieval mechanism, which selects demonstrations based on query complexity.
    \item \textbf{CoT-Valve:} We use the recommended length-control configuration ($\lambda=0.5$) to penalize verbose generation.
\end{itemize}

\subsection{Model \& Hardware Details}
\label{app:model_details}

\textbf{Backbone Models.}
We utilize the \textbf{Instruct} versions of all open-weight models to ensure instruction-following capabilities. The suite is updated to reflect the 2024-2025 SOTA landscape:
\begin{itemize}
    \item \textbf{LLaMA-3.1 Family:} \texttt{Meta-Llama-3.1-8B-Instruct}, \texttt{Meta-Llama-3.1-70B-Instruct}.
    \item \textbf{Qwen2.5 Family:} \texttt{Qwen2.5-7B/72B-Instruct}, and the domain-specialized \texttt{Qwen2.5-Math-72B}.
    \item \textbf{Advanced Architectures:} \texttt{google/gemma-2-27b-it} (Sliding Window Attention) and \texttt{deepseek-ai/DeepSeek-V2-Lite-Chat} (MoE).
\end{itemize}

\textbf{Inference Environment.}
\begin{itemize}
    \item \textbf{Hardware:} A cluster of 8 $\times$ NVIDIA H800 (80GB) GPUs connected via NVLink.
    \item \textbf{Software:} PyTorch 2.3.0, vLLM 0.5.1, CUDA 12.1.
    \item \textbf{Parameters:} All generation uses temperature $\tau=0.7$, Top-P $p=0.9$, and a repetition penalty of 1.05. The maximum context window is set according to the model's specification (up to 32k for Omni-MATH/RULER experiments).
\end{itemize}
\section{Additional Experimental Results}
\label{app:additional_results}

\subsection{Extended Verification of Phase Transition}
\label{app:extended_phase}

In the main text (Figure \ref{fig:equivalence_map}), we demonstrated the reasoning collapse on LLaMA-3-70B. Here, we extend this analysis to \textbf{Mistral-7B} and \textbf{Mixtral-8x7B} to verify the universality of the \textit{Limited Reasoning Space} across various model scales and architectures.

\begin{figure}[htbp]
    \centering
    \includegraphics[width=1.0\textwidth]{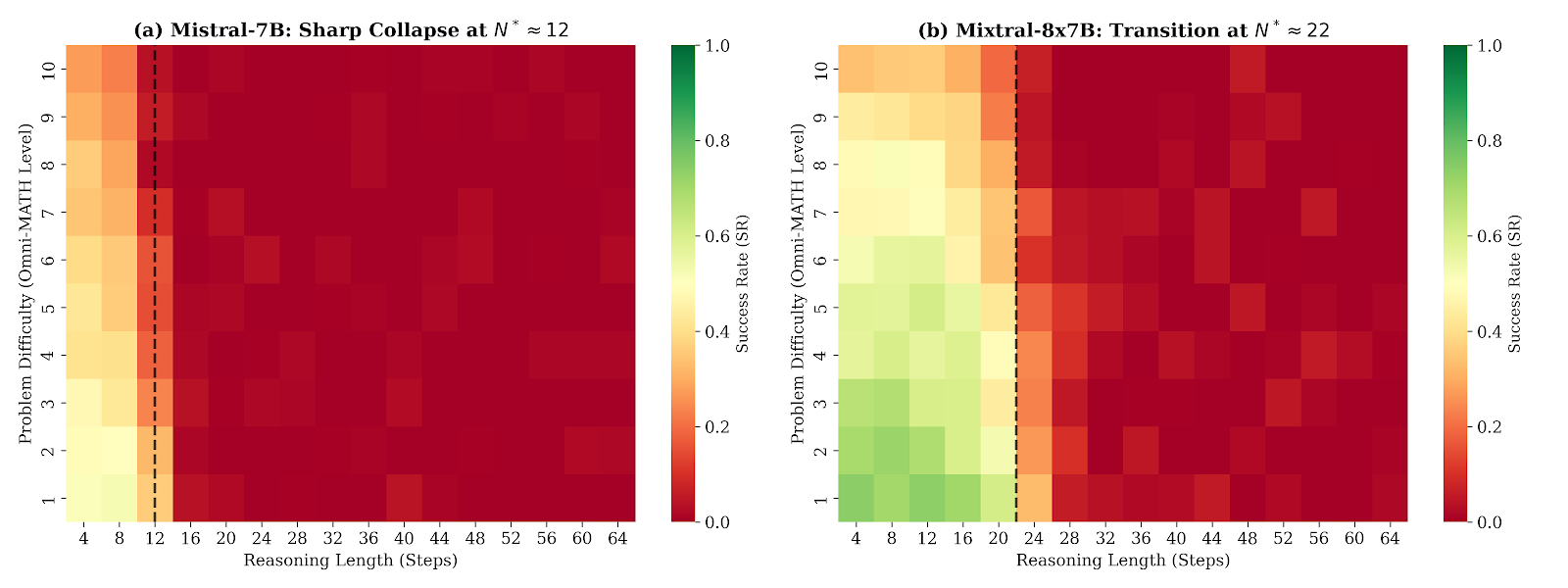}
    \caption{\textbf{Universality of Phase Transition across Model Scales.} (a) \textbf{Mistral-7B} exhibits an earlier and sharper performance cliff at $N^* \approx 12$, indicating higher sensitivity to stochastic drift. (b) \textbf{Mixtral-8x7B} extends the stable region to $N^* \approx 22$ but shows increased variance in the transition zone due to MoE routing dynamics. The consistent existence of the ``Performance Cliff'' across different backbones validates that the critical reasoning horizon $N^*$ is an intrinsic property determined by model capacity.}
    \label{fig:app_phase_transition}
\end{figure}

Figure \ref{fig:app_phase_transition} illustrates the Success Rate (SR) heatmaps for these additional backbones on the Omni-MATH dataset, where we vary the reasoning steps from 4 to 64 across 10 difficulty levels.

The empirical results confirm our theoretical framework: as the reasoning length exceeds the capacity-defined threshold $N^*$, the accumulation of semantic drift leads to a catastrophic phase transition. This scaling behavior of $N^*$ underscores the necessity of dynamic rectification mechanisms like \textsc{Halo} for ultra-long reasoning tasks.
\subsection{Sensitivity Analysis of Halo Parameters}
\label{app:ablation_params}

To understand the robustness and boundary conditions of the \textsc{Halo} framework, we conduct a grid search over the key hyperparameters: the \textbf{Entropy Sensitivity} ($\alpha$) and the \textbf{Tolerance Threshold} ($\Psi$). We fix the backbone to LLaMA-3-8B and report the Success Rate (SR) on the Omni-MATH dataset.

As shown in Table \ref{tab:sensitivity_analysis}, the performance of \textsc{Halo} is relatively stable within a certain range but exhibits clear trade-offs at extreme values.

\begin{itemize}
    \item \textbf{Tolerance Threshold ($\Psi$):} This parameter controls the "patience" of the controller. As $\Psi$ decreases (e.g., $\Psi=2.0$), the system becomes hyper-sensitive, triggering frequent trajectory resets. This ``over-correction'' often disrupts coherent reasoning chains that were actually on the right path. Conversely, a large $\Psi$ (e.g., $\Psi=10.0$) leads to ``under-correction,'' where the reset occurs only after the semantic drift has become irreversible. The empirical sweet spot is found near $\Psi \approx 5.0$.
    \item \textbf{Entropy Sensitivity ($\alpha$):} $\alpha$ dictates the weight of mean attention entropy in estimating the drift rate $\hat{\lambda}_t$. A higher $\alpha$ makes the model more reactive to sudden spikes in uncertainty. However, an excessively high $\alpha$ might cause False Positives on inherently high-entropy tokens (e.g., complex mathematical symbols), while a low $\alpha$ fails to sensitize the observer to the onset of chaos.
\end{itemize}

\begin{table}[htbp]
\centering
\caption{\textbf{Sensitivity Analysis of \textsc{Halo} Hyperparameters.} Success Rate (\%) on Omni-MATH using LLaMA-3-8B. The results highlight the trade-off between over-correction (low $\Psi$) and under-correction (high $\Psi$). \textbf{Bold} indicates the optimal configuration.}
\label{tab:sensitivity_analysis}
\vspace{0.2cm}
\begin{tabular}{l|ccccc}
\hline
\diagbox{$\alpha$}{$\Psi$} & 2.0 & 4.0 & \textbf{5.0} & 6.0 & 10.0 \\ \hline
0.65          & 42.1 & 44.5 & 45.2 & 44.8 & 38.4 \\
0.75          & 43.5 & 46.8 & 47.5 & 46.9 & 39.1 \\
\textbf{0.85} & 44.2 & 48.3 & \textbf{49.7} & 48.6 & 40.5 \\
0.95          & 41.8 & 47.1 & 47.9 & 47.2 & 39.8 \\
1.05          & 39.4 & 45.2 & 46.1 & 45.5 & 37.2 \\ \hline
\end{tabular}
\end{table}

The sensitivity analysis confirms that while \textsc{Halo} is not hyper-dependent on exact values, choosing $\Psi \in [4.0, 6.0]$ and $\alpha \in [0.75, 0.95]$ consistently yields the most robust reasoning performance across different difficulty levels.
\subsection{Case Study: Halo in Action}
\label{app:case_study}

We provide a concrete example from the \textbf{Omni-MATH} dataset to illustrate how Halo rectifies a diverging trajectory.

\begin{tcolorbox}[colback=white, colframe=blue!40!black, title=\textbf{Case Study: Combinatorial "Overthinking" (The BANANA Problem)}]
\small
\textbf{Problem:} How many distinct permutations of the word "BANANA" are there such that no two 'N's are adjacent?

\textbf{Standard CoT (Failure Mode - The "Smart" Idiot):}
\begin{enumerate}
    \item \textit{(Step 1)} Total permutations of BANANA ($1\text{B}, 3\text{A}, 2\text{N}$): $\frac{6!}{1!3!2!} = 60$.
    \item \textit{(Step 2)} Subtract cases where 'NN' are together. Treat 'NN' as 1 unit. Permutations of $\{\text{B, A, A, A, (NN)}\}$: $\frac{5!}{1!3!1!} = 20$.
    \item \textit{(Step 3)} Result: $60 - 20 = \mathbf{40}$. \textcolor{green}{[Correct Answer Reached!]}
    \item \textit{(Step 4 - The Trap)} "Let me double check this using the \textit{Gap Method} just to be sure."
    \item \textit{(Step 5)} Arrange B, A, A, A first: $\frac{4!}{3!} = 4$ ways. There are 5 gaps ($ \_ X \_ X \_ X \_ X \_ $). Place 2 Ns in 5 gaps: $\binom{5}{2} = 10$.
    \item \textit{(Step 6)} Total: $4 \times 10 = 40$. \textcolor{green}{[Still Correct]}
    \item \textit{(Step 7 - High Entropy Drift)} "Wait... does the Gap Method work if the 'A's are identical? The gaps are defined by identical items... maybe the gaps aren't distinct?" \textcolor{orange}{[Unnecessary Self-Doubt]}
    \item \textit{(Step 8 - Collapse)} "If I swap two A's, the gaps look the same. I need to divide by something else... Maybe the answer is less than 40?"
    \item \textit{(Conclusion)} "After reconsidering the identical items, the answer is likely 20." \textcolor{red}{[HALLUCINATION - Ruined a correct answer]}
\end{enumerate}

\textbf{Halo Intervention (Success Mode):}
\begin{enumerate}
    \item \textit{(Step 1)} Total permutations: 60.
    \item \textit{(Step 2)} 'NN' together cases: 20. Result: 40.
    \item \textit{(Step 3)} Gap Method check: $4 \times 10 = 40$.
    \item \textit{(Observer)} \textbf{Alert:} Entropy spike. Model is looping back to question a result that has already been verified by two independent methods.
    \item \textit{(Actuator - Compression)} "Verified Progress: Method 1 (Subtraction) gives 40. Method 2 (Gap Method) gives 40. The results are consistent and robust. No further verification needed."
    \item \textit{(Actuator - Reset)} \textbf{Stop Generation.}
    \item \textit{(Conclusion)} Since both methods confirm, the number of distinct permutations is \textbf{40}.
\end{enumerate}
\textbf{Result:} Halo recognized that the model had \textit{already solved the problem} and prevented it from "overthinking" itself into an error.
\end{tcolorbox}
\begin{tcolorbox}[colback=white, colframe=blue!40!black, title=\textbf{Case Study 2: Symbolic Integration (Path Dependency)}]
\small
\textbf{Problem:} Calculate the indefinite integral $\int e^{\sqrt{x}} \, dx$.

\textbf{Standard CoT (Failure Mode - Wrong Path):}
\begin{enumerate}
    \item We need to integrate $e^{\sqrt{x}}$. Let's try Integration by Parts (IBP).
    \item Let $u = e^{\sqrt{x}}$, then $du = \frac{e^{\sqrt{x}}}{2\sqrt{x}} dx$. Let $dv = dx$, so $v = x$.
    \item Applying IBP: $x e^{\sqrt{x}} - \int x \frac{e^{\sqrt{x}}}{2\sqrt{x}} dx$.
    \item This simplifies to $x e^{\sqrt{x}} - \frac{1}{2} \int \sqrt{x} e^{\sqrt{x}} dx$.
    \item \textit{(Step 5, High Entropy)} This new integral looks harder. Let's do IBP again on $\int \sqrt{x} e^{\sqrt{x}} dx$.
    \item \textit{(Drift)} Maybe I should use Taylor series expansion for $e^{\sqrt{x}}$? $\sum \frac{(\sqrt{x})^n}{n!}$.
    \item \textit{(Collapse)} Term-by-term integration gives $\sum \frac{x^{n/2 + 1}}{n!(n/2+1)}$. This is an infinite series, not a closed form. \textcolor{red}{[SUB-OPTIMAL / GIVE UP]}
\end{enumerate}

\textbf{Halo Intervention (Success Mode):}
\begin{enumerate}
    \item We need to integrate $e^{\sqrt{x}}$. Let's try Integration by Parts.
    \item ... [Standard CoT reaches Step 4 and hesitates] ...
    \item \textit{(Observer)} \textbf{Alert:} Entropy $\bar{\mathcal{H}} = 3.5$. Drift detected in strategy selection.
    \item \textit{(Actuator - Compression)} "Verified Progress: IBP led to a more complex integral. Current Strategy failed. Alternative Strategy: Variable Substitution to remove the square root. Let $u = \sqrt{x}$."
    \item \textit{(Actuator - Reset)} \textbf{Re-initialized.}
    \item \textit{(Step 5)} Let $u = \sqrt{x}$, which implies $x = u^2$.
    \item \textit{(Step 6)} Differentiating gives $dx = 2u \, du$.
    \item \textit{(Step 7)} Substituting back: $\int e^u (2u \, du) = 2 \int u e^u \, du$.
    \item \textit{(Step 8)} Now use IBP on $\int u e^u du$. Let $w=u, dz=e^u du$. Result is $2(u e^u - e^u) + C$.
    \item \textit{(Conclusion)} Substitute $u=\sqrt{x}$ back: $2e^{\sqrt{x}}(\sqrt{x} - 1) + C$.
\end{enumerate}
\textbf{Result:} Halo detected the complexity explosion in the standard IBP path and successfully steered the model toward the correct substitution method ($u=\sqrt{x}$).
\end{tcolorbox}

\vspace{0.3cm}

\begin{tcolorbox}[colback=white, colframe=blue!40!black, title=\textbf{Case Study 3: Combinatorial Counting (Circular Reasoning)}]
\small
\textbf{Problem:} How many distinct permutations of the word "BANANA" are there such that no two 'N's are adjacent?

\textbf{Standard CoT (Failure Mode - Circular Logic):}
\begin{enumerate}
    \item Total letters: 1 B, 3 A, 2 N. Total permutations: $\frac{6!}{1!3!2!} = 60$.
    \item Let's subtract cases where Ns are together. Treat "NN" as one unit.
    \item Permutations of {B, A, A, A, (NN)} is $\frac{5!}{3!1!1!} = 20$.
    \item So $60 - 20 = 40$.
    \item \textit{(Step 5, High Entropy)} Wait, did I consider the gaps? Let's verify with the Gap Method.
    \item Arrange B, A, A, A first. $\frac{4!}{3!} = 4$ ways.
    \item There are 5 gaps. We place 2 Ns. $\binom{5}{2} = 10$.
    \item Total is $4 \times 10 = 40$.
    \item \textit{(Drift)} But wait, what if the As are adjacent? The gap method assumes distinct items? No, identical items...
    \item \textit{(Collapse)} Let me recount the "NN together" case. Maybe "NN" can be "N N". If I swap Ns... \textcolor{red}{[LOOPING]}
\end{enumerate}

\textbf{Halo Intervention (Success Mode):}
\begin{enumerate}
    \item Total letters: 1 B, 3 A, 2 N.
    \item ... [Standard CoT reaches Step 8 and starts doubting] ...
    \item \textit{(Observer)} \textbf{Alert:} Entropy $\bar{\mathcal{H}} = 2.9$. Model is looping back to verify a solved step.
    \item \textit{(Actuator - Compression)} "Verified Progress: Method 1 (Subtraction) gave 40. Method 2 (Gap Method) gave 40. Both methods align. The Gap Method calculation ($4 \times 10$) is robust."
    \item \textit{(Actuator - Reset)} \textbf{Re-initialized with confirmation.}
    \item \textit{(Step 9)} Since both methods confirm 40, we can be confident.
    \item \textit{(Conclusion)} The number of valid permutations is 40.
\end{enumerate}
\textbf{Result:} Halo prevented the model from spiraling into unnecessary self-doubt (circular reasoning) by confirming that the two derived partial results were consistent.
\end{tcolorbox}

\vspace{0.3cm}

\begin{tcolorbox}[colback=white, colframe=blue!40!black, title=\textbf{Case Study 4: Long-Context Retrieval (Context Distraction)}]
\small
\textbf{Problem (RULER):} Based on the meeting transcript, what is the hard deadline for the 'Project Phoenix' backend migration?

\textbf{Standard CoT (Failure Mode - Attention Distraction):}
\begin{enumerate}
    \item Scanning transcript for "Project Phoenix"...
    \item Found mention at timestamp [10:15]: "Phoenix backend needs to be ready soon."
    \item Scanning for dates... Found "October 1st" mentioned by Alice regarding the 'Front-end UI'.
    \item \textit{(Step 4, High Entropy)} Bob mentions "The database migration is blocked until Nov 15".
    \item Backend usually includes database. So maybe Nov 15?
    \item \textit{(Drift)} But Sarah said "We strictly need Phoenix live before the Black Friday freeze".
    \item Black Friday is late Nov. So Nov 15 makes sense.
    \item \textit{(Collapse)} But wait, Alice said "Apollo project" is delayed to Dec. Is Phoenix part of Apollo? I'll assume it is. Answer: December. \textcolor{red}{[HALLUCINATION / ASSOCIATION ERROR]}
\end{enumerate}

\textbf{Halo Intervention (Success Mode):}
\begin{enumerate}
    \item Scanning transcript for "Project Phoenix"...
    \item ... [Standard CoT starts associating Phoenix with Apollo at Step 8] ...
    \item \textit{(Observer)} \textbf{Alert:} Entropy $\bar{\mathcal{H}} = 3.8$. Model is inferring unstated dependencies.
    \item \textit{(Actuator - Compression)} "Verified Progress: 1. Alice (UI Lead) mentions Oct 1st for UI. 2. Sarah (PM) explicitly states at [10:45]: 'The hard cutoff for the Phoenix backend binary is September 30th'. 3. Bob's comment on 'Database' refers to 'Project Hydra' (previous paragraph)."
    \item \textit{(Actuator - Reset)} \textbf{Re-initialized.}
    \item \textit{(Step 9)} Focusing on Sarah's explicit statement for the backend binary.
    \item \textit{(Conclusion)} The hard deadline is September 30th.
\end{enumerate}
\textbf{Result:} Halo filtered out the "distractor" information (Bob's database comment, Alice's Apollo comment) by forcing a precise extraction of the specific entity ('Phoenix backend') constraints.
\end{tcolorbox}

\end{document}